\documentclass[10pt,twocolumn,letterpaper]{article}

\usepackage{cvpr}
\usepackage{times}
\usepackage{epsfig}
\usepackage{graphicx}
\usepackage{amsmath}
\usepackage{amssymb}
\usepackage{epstopdf}
\usepackage{bbm}

\DeclareMathOperator*{\argmax}{arg\,max}
\newcommand{\norm}[1]{\left\lVert#1\right\rVert}

\usepackage[pagebackref=true,breaklinks=true,letterpaper=true,colorlinks,bookmarks=false]{hyperref}

 \cvprfinalcopy 

\def\cvprPaperID{548} 

\ifcvprfinal\fi
\begin{document}

\title{SHOE: Supervised Hashing with Output Embeddings}

\author{Sravanthi Bondugula, Varun Manjunatha, Larry S. Davis, David Doermann\\
{\tt\small sravb@cs.umd.edu, varunm@cs.umd.edu, lsd@umiacs.umd.edu, doermann@umd.edu}\\
University of Maryland College Park, MD\\
}

\maketitle

\begin{abstract}
We present a supervised binary encoding scheme for image retrieval that learns projections by taking into account similarity between classes obtained from output embeddings. Our motivation is that binary hash codes learned in this way improve both the visual quality of retrieval results and existing supervised hashing schemes. We employ a sequential greedy optimization that learns relationship aware projections by minimizing the difference between inner products of binary codes and output embedding vectors. We develop a joint optimization framework to learn projections which improve the accuracy of supervised hashing over the current state of the art with respect to standard and sibling evaluation metrics. We further boost performance by applying the supervised dimensionality reduction technique on kernelized input CNN features. Experiments are performed on three datasets: CUB-2011, SUN-Attribute and ImageNet ILSVRC 2010. As a by-product of our method, we show that using a simple k-nn pooling classifier with our discriminative codes improves over the complex classification models on fine grained datasets like CUB and offer an impressive compression ratio of 1024 on CNN features.
\end{abstract}

\section{Introduction}
Given a database of images, image retrieval is the problem of returning images from the database that are most similar to a query. Performing image retrieval on databases with billions of images is challenging due to the linear time complexity of nearest neighbor retrieval algorithms. Image hashing\cite{lsh,klsh,sh,ksh, fasthash, mlh} addresses this problem by obtaining similarity preserving binary codes which represent high dimensional floating point image descriptors, and offer efficient storage and scalable retrieval with sub-linear search times. These binary hash-codes can be learned in unsupervised or supervised settings. Unsupervised hashing algorithms map nearby points in a metric space to similar binary codes. Supervised hashing algorithms try to preserve semantic label information in the Hamming space. Images that belong to the same class are mapped to similar binary codes.

\begin{figure}[]
\begin{center}
   \includegraphics[height=3.5cm,width=8.5cm]{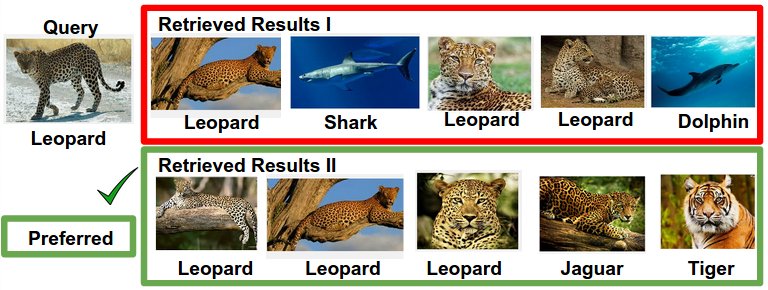}
\end{center}
   \caption{We prefer results II over I because they tend to retrieve images of classes(jaguar and tiger) related to the class label of the query(leopard), rather than unrelated classes(sharks and dolphins).}
\label{fig:toyexample}
\end{figure}

In this work, we develop a new approach to supervised hashing, which we motivate with the following example (Figure~\ref{fig:toyexample}). Consider an image retrieval problem that involves a database of animals. A user provides the query image of a leopard. Now consider the following three scenarios : 
\begin{enumerate}
\item If the retrieval algorithm returns images of leopards, we can deem the result to be absolutely satisfactory. 
\item If the retrieval algorithm returns images of dolphins, whales or sharks, we consider the results to be absolutely unsatisfactory because not only are dolphins not leopards, they do not look anything at all like leopards. 
\item If the retrieval algorithm returns the image of a jaguar or a tiger, we would be reasonably satisfied with the results. Although a jaguar is not the same as a leopard, it does look remarkably similar to one - it is a large carnivorous cat with spots and a tawny yellow coat. 
\end{enumerate}

In the above example, leopards, dolphins, whales, sharks, jaguars and tigers all belong to different categories. However, some of these categories are more closely related to each other than to other categories. Animals which fall under the ``big cat'' (\emph{Panthera}) genus are related to each other, as are the large aquatic vertebrates like dolphins, sharks and whales. We designate the related categories as ``siblings''. To study the relationships between categories, Weinberger \emph{et al.}\cite{largemargin} suggested the concept of ``output embeddings'' - vector representations of category information in Euclidean space. There has been extensive work  on ``input embeddings'', which are vector-space representations of images\cite{bow,fv,krizhevsky}, but less work has been done on output embeddings, which map similar category labels to similar vectors in Euclidean space. In an output embedding space of animals, we would expect to have embeddings for labels so that, chimpanzees, orangutans and gorillas are near each other as are leopards, cheetas, tigers and jaguars. 

\begin{figure}[t]
\begin{center}
   \includegraphics[height=4.5cm,width=8.5cm]{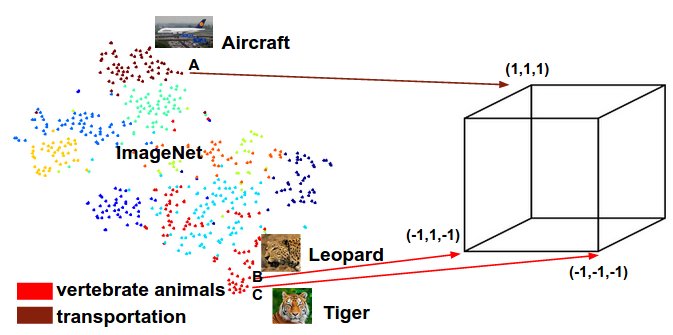}
\end{center}
   \caption{We perform k-means clustering on Word2Vec embeddings \cite{mikolov} of ImageNet classes. The principle behind SHOE is that images belonging to related classes (like leopard or tiger, which are nearby in the output embedding space) are mapped to nearby binary codes (represented by points on a binary hypercube). Images belonging to unrelated classes (like leopards and aircraft) are mapped to distant binary codes. This figure was created using \cite{tsne} and is for illustrative purposes.} 
\label{fig:imagenetvis}
\end{figure}

Our method, which uses output embeddings to construct hash codes in a supervised framework is called SHOE: Supervised Hashing with Output Embeddings (Figure~\ref{fig:imagenetvis}). Our motivation for doing this is two-fold. First, it is our belief (validated experimentally) that we can construct better binary codes for a particular class given knowledge of its sibling classes. Secondly, if our algorithm is unable to retrieve images of the same class, it will try to retrieve images of sibling classes, rather than images of unrelated classes. The assumption here is that if images of the same class as the query cannot be retrieved, images of sibling classes are more useful to the user than images of unrelated classes. We perform extensive retrieval experiments on the Caltech-UCSD Birds(CUB) dataset\cite{cub}, SUN Attribute Dataset\cite{sun} and ImageNet\cite{ilsvrc2010}. Our hash-codes can also be used to do classification, and we report accuracy on the CUB dataset using a nearest-neighbor classifier which is better than R-CNN and its variants \cite{rcnn, partrcnn}.

\noindent
The contributions of our paper are as follows : 
\begin{enumerate}
\item To the best of our knowledge, our approach is the first to introduce the problem of learning supervised hash functions using output embeddings. 
\item We propose a joint learning method to solve the above problem, and perform retrieval and classification experiments to experimentally validate our method. 
\item We propose two new evaluation criteria - ``sibling metrics'' and ``weighted sibling metrics'', for gauging the efficacy of our method.
\item We significantly boost retrieval and classification performance by applying Canonical Correlation Analysis \cite{cca} on supervised features, and learn hash functions using output embeddings on these features.
\end{enumerate}

The remainder of this paper is arranged as follows. Section ~\ref{sec:related} describes related work. In Section ~\ref{sec:method} we describe our hashing framework, and carry out experiments in Sections ~\ref{sec:experiments} and ~\ref{sec:classification}. We conclude in Section ~\ref{sec:conclusion}.

\section{Related Work}
\label{sec:related}

Work on image hashing can be divided into unsupervised and supervised methods. For the purpose of brevity, we only consider supervised methods. Supervised Hashing algorithms are based on the objective function of minimizing the difference between hamming distances and similarity of pairs of data points. Supervised Hashing with Kernels(KSH)\cite{ksh} uses class labels to determine the similarity. Points are considered `similar' (value `1') if they belong to the same class and `dissimilar` (value `-1') otherwise.  They utilize a simplified objective function using the relation between the hamming distance and inner products of the binary codes. A sequential greedy optimization is adapted to obtain supervised projections. FastHash \cite{fasthash} also uses a KSH objective function but employs decision trees as hash functions and utilizes a GraphCut based method for binary code inference. Minimal loss hashing \cite{mlh} uses a structured SVM framework \cite{ssvm} to generate binary codes with an online learning algorithm. 

All these methods except KSH categorize the input pairs to be either similar or dissimilar. KSH allows a similarity 0 for related pairs, but the authors only use it to define metric neighbors and not for semantic neighbors. FastHash entirely ignores the related pairs, as it weighs the KSH loss function by the absolute value of the label. Also, their work does not support a similarity value other than 1 and -1, as it violates the submodularity property - a crucial property required to solve the problem in parts. To the best of our knowledge, we are the first to use similarity information of sibling classes in a supervised hashing framework. Our work is also different from others as we compute similarity from output embeddings and use a joint learning framework to learn the sibling similarity. We now discuss related work about output embeddings.



Output embeddings can be defined as vector representations of class labels. These are divided into two types: data-independent and data-dependent. Some of the data-independent embeddings include \cite{langford, cub, awa, sun, taxonomy}. Langford \emph{et al.} \cite{langford} constructed output embeddings randomly from the rows of a Hadamard matrix where each embedding was a random vector of 1 or -1. Embeddings constructed from side information about classes such as attributes, or a Linnean hierarchy are available with datasets such as \cite{cub, awa, sun}. In WSABIE\cite{wsabie}, the authors jointly learn the input embeddings and the output embeddings to maximize classification accuracy in a structural SVM setting. Akata \emph{et al.}\cite{akata} uses the WSABIE framework to learn fine grained classification models by mapping the output embeddings to attributes, taxonomies and their combination.  

The binary hash-codes that our method, SHOE, learns on the CUB and SUN datasets use attributes as output embeddings, just like \cite{akata}. For ILSVRC2010 experiments, we use a taxonomy derived embeddings similar to \cite{taxonomy}. In a taxonomy embedding, a binary output embedding vector is obtained, where each node in the class hierarchy and its ancestors are represented as 1 while non-ancestors are represented as 0. Deng \emph{et al.} \cite{deng} show that classification that takes hierarchies into account can be informative. Mikolov  \emph{et al.} \cite{mikolov} use a skip-gram architecture trained on a large text corpus to learn output embeddings for words and short phrases. These Word2Vec embeddings are used by \cite{devise} for large scale image classification and zero-shot learning. Finally, output embeddings can even be learned from the data. For example, \cite{costa} exploits co-occurences of visual concepts to learn classifiers for unseen labels using known classifiers. All these methods use output embeddings for classification and zero shot learning, but none have used them to learn binary codes for retrieval.

\section{Method}
\label{sec:method}
\begin{figure*}
\begin{center}
\begin{minipage}[t]{.68\linewidth}
\vspace{0pt}
\centering
 \includegraphics[width=0.3\linewidth]{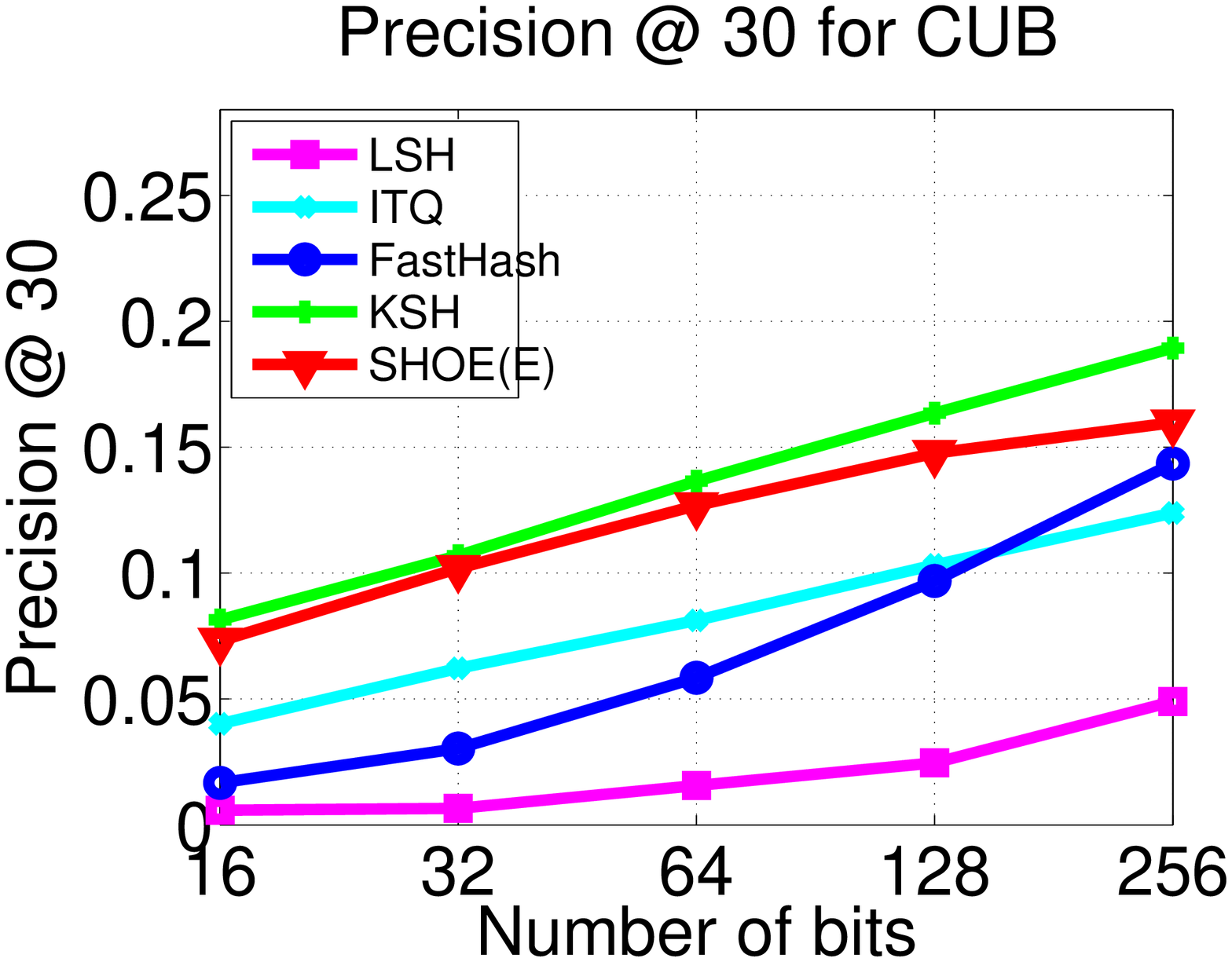}
 \includegraphics[width=0.3\linewidth]{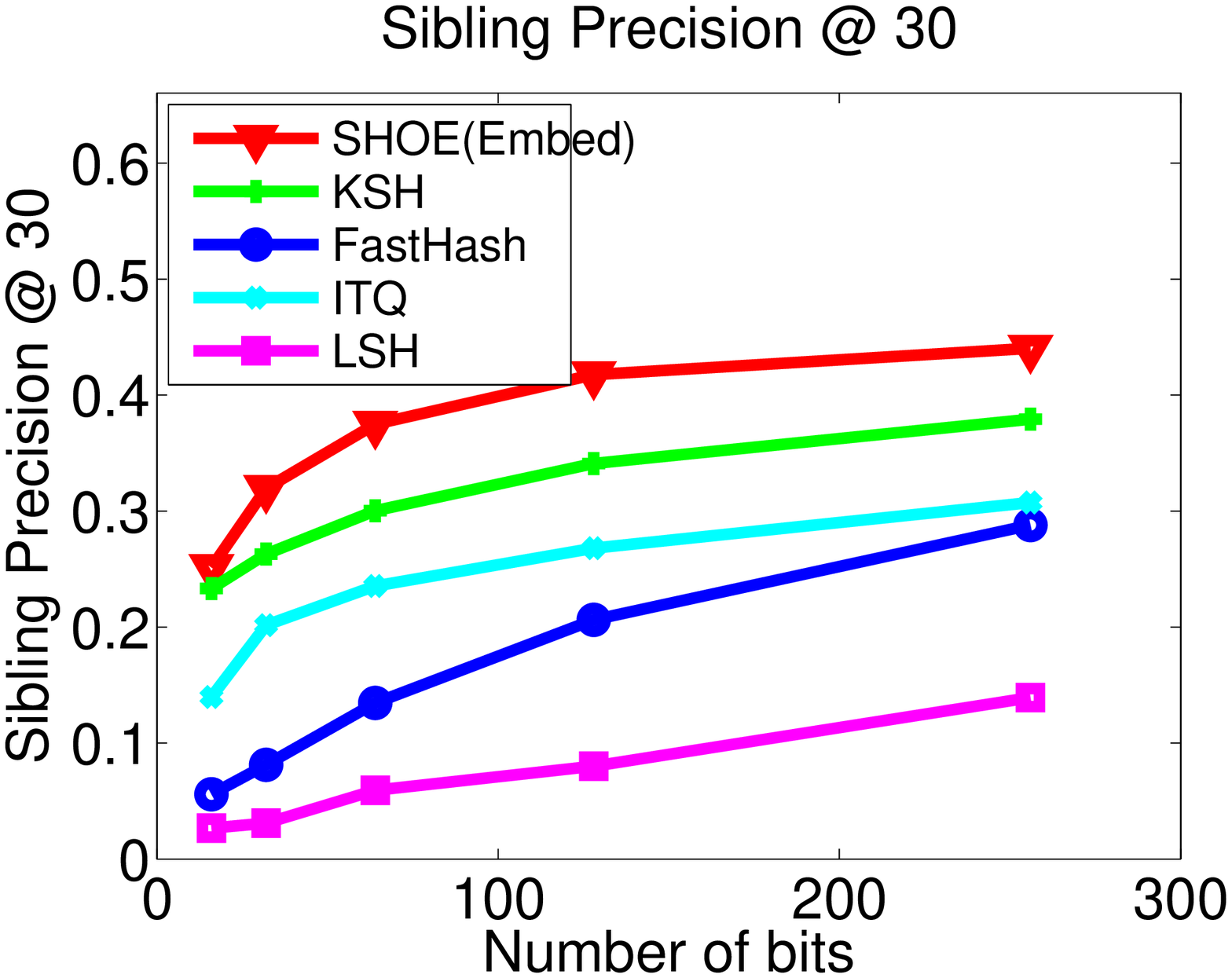}
 \includegraphics[width=0.3\linewidth]{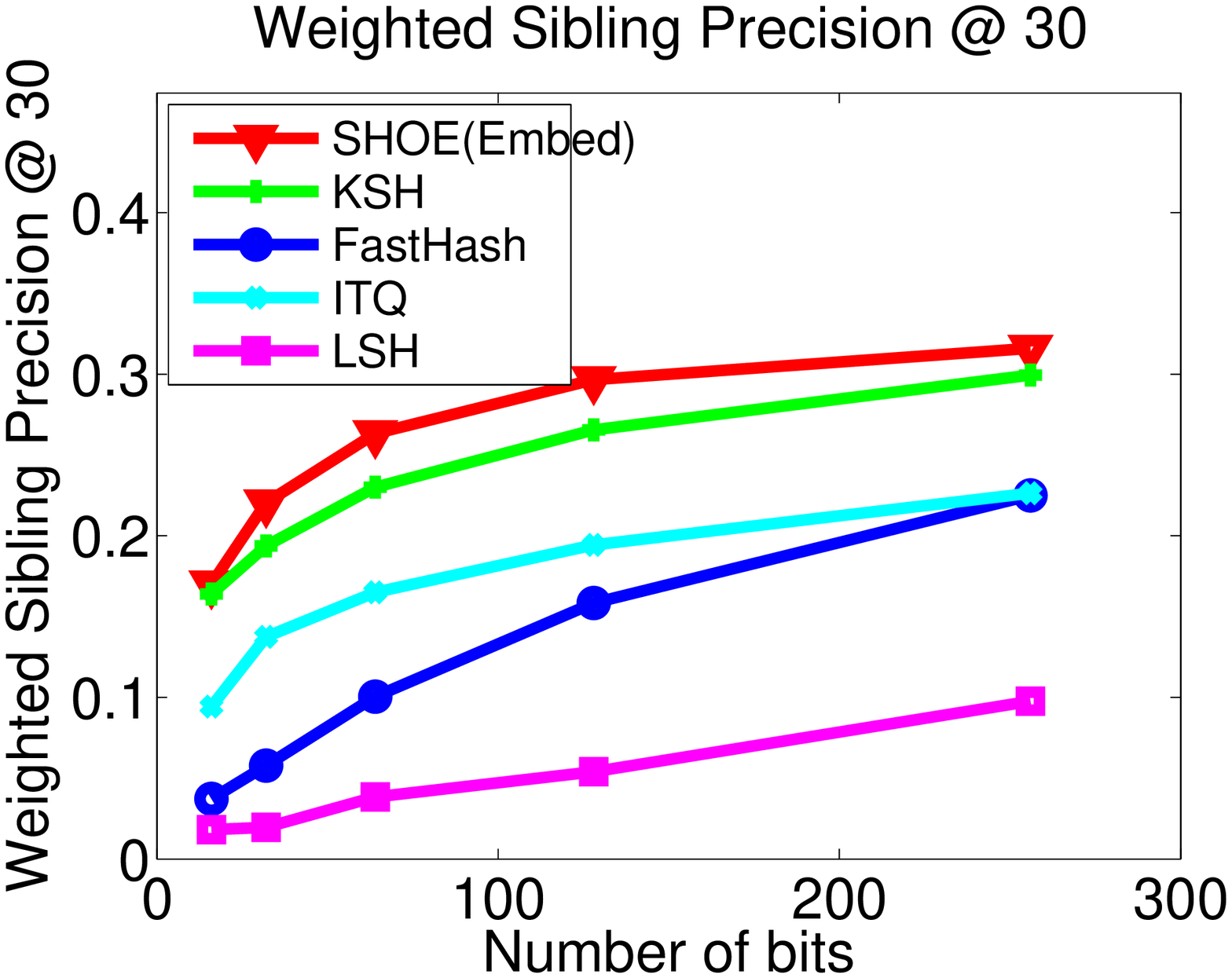}
\end{minipage}%
\begin{minipage}[t]{0.32\linewidth}
\vspace{0pt}
\centering
\small
\begin{tabular}{|c|c|c|c|}

\hline
Method & mAP & $Sib_{mAP}$ & $Sib_{mAP}^w$  \\
\hline
SHOE(E) &  0.111 & \textbf{0.250} & \textbf{0.174} \\ \hline
KSH & \textbf{0.113}  & 0.133 & 0.108 \\ \hline
FastHash & 0.045  & 0.062 & 0.047 \\ \hline
ITQ & 0.060  & 0.119 & 0.084 \\ \hline
LSH & 0.013  & 0.044 & 0.028 \\ \hline
\end{tabular}
\end{minipage}
\normalsize
\end{center}
\caption{ Retrieval on CUB dataset comparing our method SHOE(E) with the state-of-the art hashing techniques. The above plots report precision, sibling precision and weighted sibling precision for top 5 sibling classes for bits $c=\{16, 32, 64, 128, 256\}$. The table reports mAP, Sibling and Weighted Sibling mAP for $c=64$ bits. }
\label{fig:cub2000compare}
\end{figure*}

\subsection{Preliminaries}
Given a training set $M = \{(x_1, y_1), .... , (x_N, y_N) \} $ of $N$ (image,label) pairs with $x_i \in \mathcal{X}$ and $y_i \in \mathcal{Y}$, let $\phi: \mathcal{X} \rightarrow \bar{\mathcal{X}} \in \mathcal{R}^d$ be the input embedding function and $\psi: \mathcal{Y} \rightarrow \bar{\mathcal{Y}} \in \mathcal{R}^e$ be the output embedding function. We wish to learn binary codes $b_i, b_j$ of length $c$ (i.e., $b_i \in \{-1, 1\}^c$) such that for pairs of training images, the Hamming distance between the codes preserve the distance between their class labels (given by their corresponding output embedding vectors). In other words, for a given query image, retrieved results of sibling(unrelated) classes ought to be ranked higher(lower). To this end, we obtain the following objective function:

\small
\begin{equation} \label{eq:distobj}
   \min O(b) = \sum_{i=1}^N\sum_{j=1}^N (d_H(b_i, b_j) - d_E(\psi(y_i), \psi(y_j)))^2.
\end{equation}
\normalsize
where $d_H(b_i, b_j)$ is the Hamming distance between binary codes $b_i$ and $b_j$ and $d_E$ is the Euclidean distance between the output embedding vectors $\psi(y_i)$ and $\psi(y_j)$.

For an input image $x$ with input embedding vector $\phi(x)$, we obtain binary code $b$ of length $c$ bits. Each bit is computed using a hash function $h_l(x)$ that takes the form : 

\begin{equation}
	h_l(x) = sgn(w_l \phi(x)), w_l \in \mathcal{R}^d.
\end{equation}
To learn $c$ such hash functions $\mathcal{H} = \{h_l | l = 1, \ldots ,c\}$, we learn $c$ projection vectors $W = [w_1, w_2  \ldots,w_c]$, which we compactly write as $H(x) = sgn(W\phi(x))$, $W \in \mathcal{R}^{c\times d}$. Without loss of generality, we can assume that $\phi(x)$ is a mean-centered feature. This ensures we obtain compact codes by satisfying the balanced property of hashing (i.e, each bit fires approximately 50\% of the time)\cite{sh}. $\phi(x)$ can be an input embedding that maps images to features in either kernelized or unkernelized forms. 

Solving the optimization problem in Equation (1) is not straightforward, so we utilize the relation between inner product of binary codes and Hamming distances \cite{mdsh, ksh}, given as $2d_H(b_i, b_j) = c - b_i^T b_j$, where $b_i^T b_j = \sum_{l=1}^c h_l(\phi(x_i)) h_l(\phi(x_j))$ is the inner product of the binary codes $b_i$ and $b_j$. Note that the inner product of binary codes lies between $-c$ and $+c$, while the Hamming distance ranges from $0$ to $c$, where the distance between the nearest neighbors is $0$ and between the farthest neighbors is $c$ . By unit normalizing the output embedding vectors, $\|\bar{\psi}(y)\| = 1$, we exploit the relationship between Euclidean distances and the dot products of normalized vectors, given as $d_E(\bar{\psi}(y_i), \bar{\psi}(y_j)))^2 = 2 - 2\bar{\psi}(y_i)^T\bar{\psi}(y_j)$, and obtain the following objective function:
\small
\begin{equation}\label{eq:dotprodobj}
   \min_{\substack{\mathcal{H}}} \sum_{i=1}^N\sum_{j=1}^N ( \frac{1}{c}\sum_{l=1}^c h_l(\phi(x_i)) h_l(\phi(x_j)) - \bar{\psi}(y_i)^T\bar{\psi}(y_j))^2
\end{equation}
\normalsize

Let $o_{ij} = \bar{\psi}(y_i)^T\bar{\psi}(y_j)$ and as a consequence of the unit normalization of $\bar{\psi}(y_i)$, $-1 \leq o_{ij} \leq 1$, which implies that the similarity between same classes is \textbf{1} and similarity between different classes is as low as \textbf{-1}. The objective ensures that the learned binary codes preserve the similarity between output embeddings, which is required for supervised hashing and our goal of ranking related neighbors before farthest neighbors.

This is similar to the KSH\cite{ksh} objective function, except that KSH assumes that $o_{ij}$ takes only values $1$($-1$) for similar(dissimilar) pairs defined with semantic information. Their work also accomodates the definition of the $o_{ij}=0$ for related pairs but only for metric neighbors. Our work is different from theirs, as we emphasize the learning of binary codes that preserve the similarity between the classes, whereas $o_{ij}$ captures the similarity between the classes. Regardless of the definition of $o_{ij}$, our optimization is similar, and so we employ a similar sequential greedy optimization for minimizing $O(b)$. We refer the reader to \cite{ksh} for further details.


\subsection{Evaluation Criteria}
\label{sec:sibeval}
Standard metrics like precision, recall and mAP defined for semantic neighbors are not sufficient to evaluate the retrieval of the sibling class images. To measure this, we define \emph{sibling precision}, \emph{sibling recall} and \emph{sibling average precision} metrics. Let $R_y:(y_i, y_j) \rightarrow rank$ return the rank of class $y_j$ for a query class $y_i$, $0 \leq rank \leq L$, where $L$ is the number of classes. Note that, $R_y(y, y)=0$ for the same class. The ranking $R_y$ is computed by sorting the distance between the output embedding vectors $\psi(y)$ and $\psi(y^*)$, $y^* \in \mathcal{Y} \setminus y$.  We obtain the weight of the sibling class used for evaluation using the following functions for Sibling($Sib_m$) and Weighted Sibling($Sib^w_m$) metrics:
\begin{equation*}
Sib_m: (y_i, y_j, R_y) \rightarrow \mathbb{I}(R_y(y_i, y_j)) \leq m)
\end{equation*}
\vspace{-0.3cm}
\begin{equation*} 
Sib^w_m:(y_i,y_j, R_y) \rightarrow \frac{m-R_y(y_i,y_j)}{m}*\mathbb{I}(rank \leq m)
\end{equation*}
where, $m$ is the number of related classes for each query class and $\mathbb{I}(.)$ is the Indicator function that returns $0$ when $rank>m$.  The Sibling and Weighted Sibling precision@k, recall@k and mAP is defined as:



\vspace{-0.3cm}
\begin{equation}
   s_{precision_q}^{w}@k = \frac{\sum_{l=1}^{k} Sib^w_m(y_q, y_{q_l}, R_y)} {k}
\end{equation}
\vspace{-0.3cm}
\begin{equation}
   s_{recall_q}^{w}@k = \frac{\sum_{l=1}^{k} Sib^w_m(y_q, y_{q_l}, R_y)}{\sum_{p=1}^{L} \mathcal{N}_l * \mathbb{I}(Sib^w_m(y_q, y_p, R))}
\end{equation}
\vspace{-0.3cm}
\begin{equation}
   s_{AP_q}^{w} = \sum_{k=1}^{N} s_{precision_q}@k \times \triangle s_{recall_q}@k
\end{equation}
\vspace{-0.3cm}
\begin{equation}
   s_{mAP}^{w} = \frac{\sum_{q=1}^{Q} s_{AP_q}}{Q}
\end{equation}
\vspace{-0.3cm}

In the above equations, $y_{q_l}$ refers to the class of the $l^{th}$ retrieved image.

\subsection{Preliminary Experiments}
\label{sec:embedexp}
We evaluate the proposed hashing scheme that takes into account structure of the related classes with the following performance metrics: Precision@k, mAP and their sibling versions previously defined. 
\begin{itemize}
\item{\textbf{\textit{Datasets:}}} To test our approach, we use datasets that contain information about class structure. The CUB-2011 dataset \cite{cub} contains 200 fine-grained bird categories with 312 attributes and is well suited for our purpose. In this dataset, although both binary and continuous real-valued attributes are available, we use only the mean-centered real-valued attributes as output embeddings $\psi(y)$. We obtain ranking $R_y$ for each class $y$ based on these attribute embeddings. There are 5994 train and 5774 test images in the dataset. We select a subset of the dataset of size $2000$ for training, use the whole train set for retrieval and all test images as queries.  Ground truth for a query is defined label-wise and each query class has approximately 30 same class neighbors in the retrieval set. For input embeddings, we extract state-of-the-art 4096 dimesional Convolution Neural Network (CNN) features from the fc7 layer for each image using the Caffe Deep Learning library developed by \cite{caffe}. We kernelize the CNN features which take the form :$\sum_{i=1}^{p} \kappa(x,x_i)$ where $\kappa$ is a radial basis kernel, and $p$ is the cardinality of a subset of training sample, designated as ``anchor points''. We refer to these as CNN+K features and they are inherently mean-centered.

\item{\textbf{\textit{Comparison methods:}}} We compare our method with raw output embeddings:SHOE(E), with the following supervised and unsupervised hashing schemes: KSH\cite{ksh}, FastHash\cite{fasthash}, ITQ\cite{itq} and LSH\cite{lsh}. We use their publicly available implementations and set the parameters to obtain the best performance. It is important to note that none of these methods utilize the distribution of class labels in the output embedding space. The closest comparison would be to use KSH, setting the similarity of semantic class neighbors to 0 value. For this purpose, we obtain top-$m$ related pairs for each class using $R_y$ and set the similarity to $0$ for related pairs. To evaluate the unsupervised ITQ and LSH, we zero-center the data and apply PCA to learn the projections.  We use CNN+K features with $p=300$ for evaluating SHOE(E) and KSH since we learn linear projections in these methods, unlike FastHash which learns non-linear decision trees on linear features.
\item{\textbf{\textit{Results:}}}  Figure~\ref{fig:cub2000compare} shows the precision@30, sibling and weighted sibling precision@30 plots for 5 related classes, i.e. $m=6$(+1 for the same class) by encoding the input embeddings to bits of length $c=\{16, 32, 64, 128, 256\}$. The table in Figure~\ref{fig:cub2000compare} shows the recall and mAP and its sibling variants for 64 bits. We observe that SHOE(E) does better than baselines for both sibling and weighted sibling precision metrics for top-30 retrieved neigbhors for all bit lengths, but there is a loss in precision compared to the KSH method. In their paper, FastHash\cite{fasthash} shows better performance compared to KSH. However, it does not perform well here because of the large number of classes and few training samples available per class. 

\end{itemize}

\subsection{Analysis}
\begin{figure}[hb]
\begin{center}
 \includegraphics[width=0.48\linewidth]{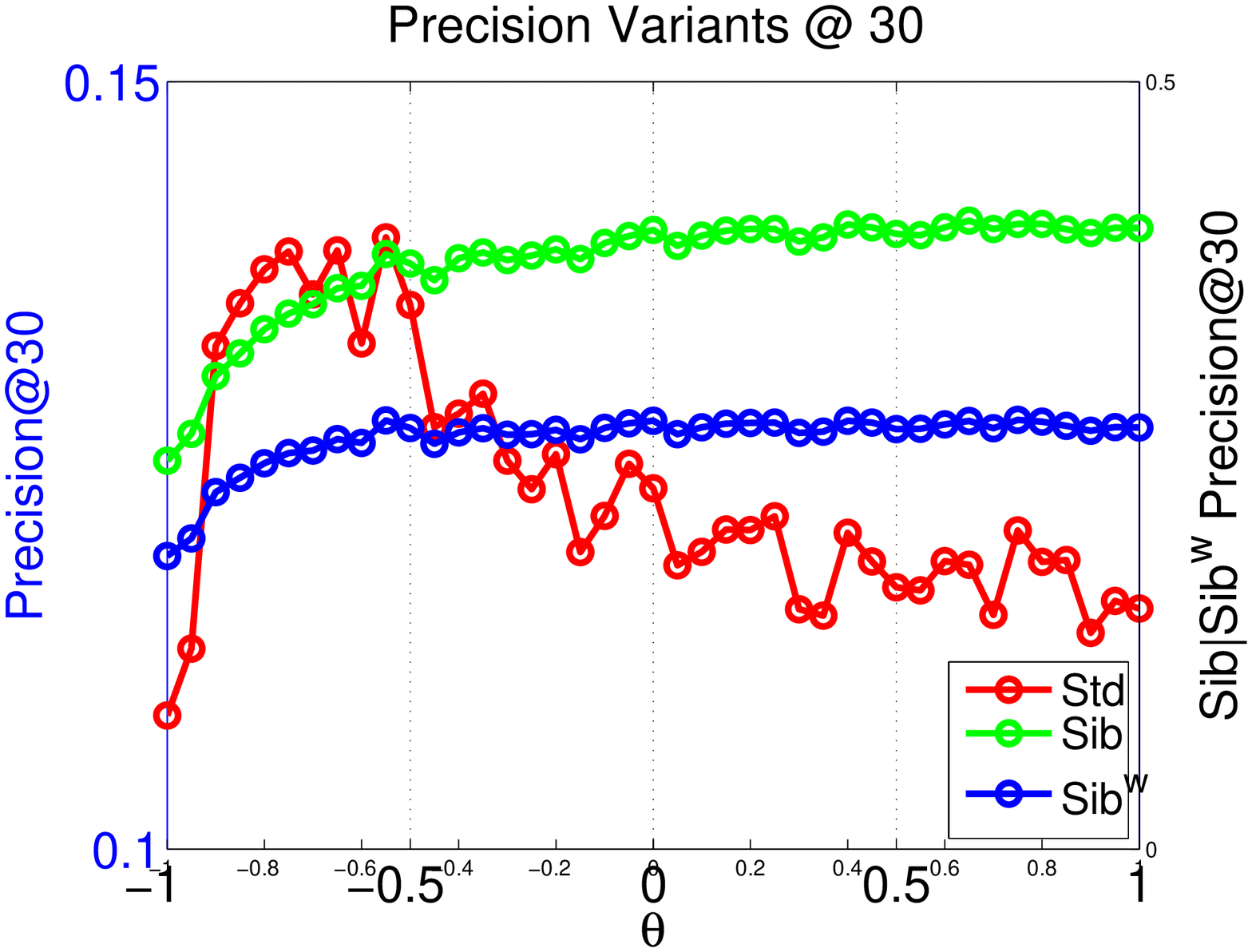}
 \includegraphics[width=0.48\linewidth]{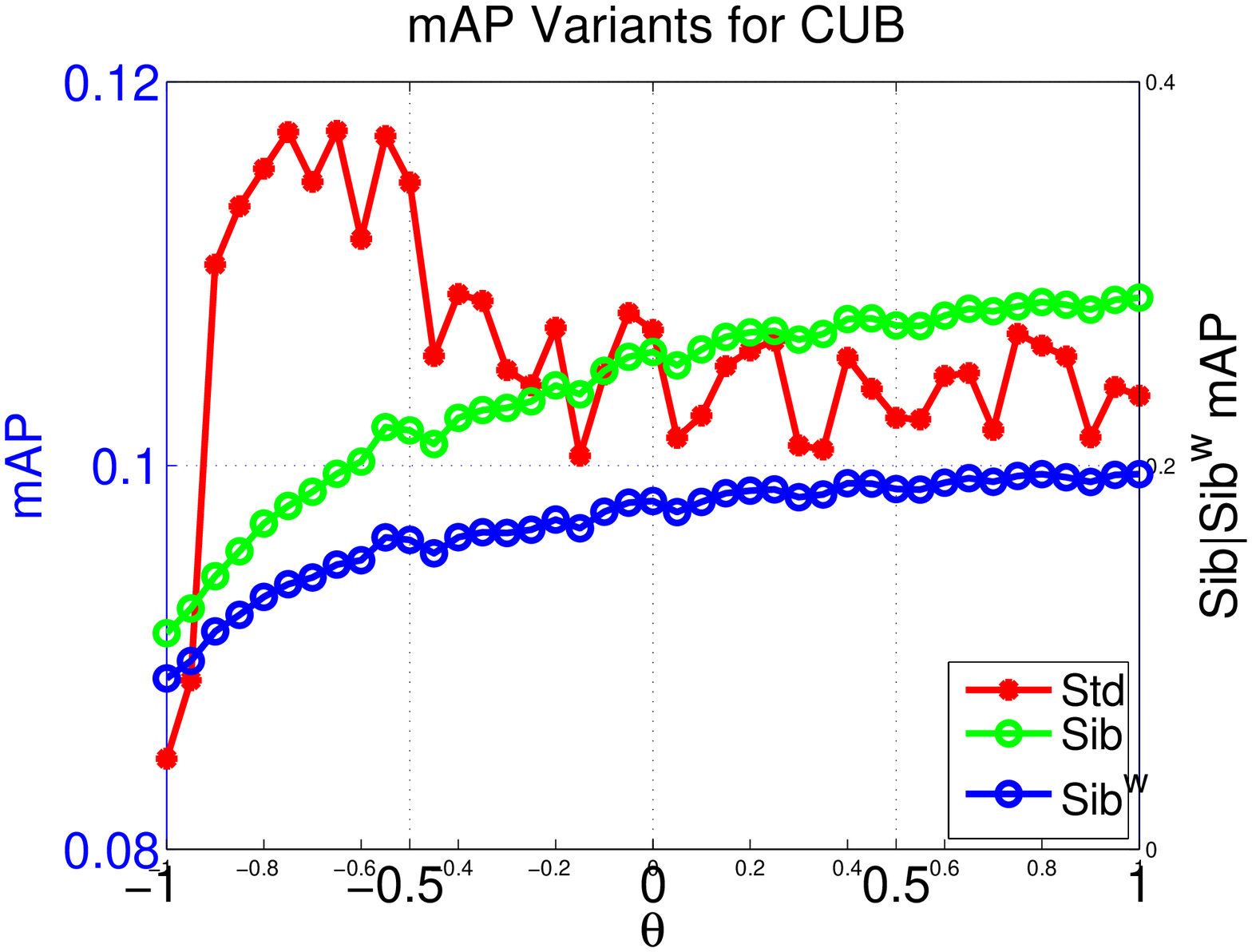}
\end{center}
   \caption{ Retrieval on CUB dataset evaluating the performance of our method(SHOE) for varying $\theta$ values and $p=1000$. The left and right y-axis show the standard metrics and sibling metrics respectively.}
\label{fig:cub2000analysisr}
\end{figure}
Contrary to our expectation that related class information improves both standard and sibling performance metrics, experiments in ~\ref{sec:embedexp} show that using similarity directly from output embeddings actually reduces the performance for same classes, while improving it for sibling classes. To analyse this, we obtained top-$m$ related classes using ranking $R_y$ and assigned a constant $o_{ij} = \theta$ value (previously, in Equation\eqref{eq:dotprodobj}, we had defined $o_{ij} = \bar{\psi}(y_i)^T\bar{\psi}(y_j)$). $\theta$ measures the similarity between a class and a related class. Figure ~\ref{fig:cub2000analysisr} show the performance of our method(SHOE) for varying $\theta$ values for 64 bits on CUB-2011 dataset. Results reveal that, when a fixed similarity is used, we actually gain performance from the sibling class training examples, and this gain is maximized for negative values of $\theta$, i.e $-1< \theta < 0$.  The intuition behind this is: when $\theta$ is close to 1, the learned hash-code would not discriminate well enough between identical classes and sibling classes. For instance, in our ``database of animals'' example, we would learn hash-codes that nearly equate leopards with jaguars, which is not what we desire. On the other hand, when $\theta$ is close to -1, a hash-code for a leopard image will be learned mostly from other training images of leopards, but with slight consideration towards training images of its sibling classes. When $\theta$ is assigned -1, the sibling classes aren't considered at all, so our method becomes identical to KSH\cite{ksh}. We are now interested in learning $\theta$ simultaneously with the hash functions during the training phase.

\subsection{SHOE Revisited}
\label{subsec:ourmethod}
We observe that the objective function that we want to minimize in Equation~\eqref{eq:dotprodobj} can be split into three parts - for identical classes, sibling classes and unrelated classes, respectively. We also observe from the preceding analysis that precision and recall metrics improve for negative values of $\theta$. Therefore, we add regularizer term $\lambda\norm{\theta + 1}^2$ to the objective function, which becomes small when $\theta$ lies close to -1. For easier notation, we denote $h_{l_i} = h_l(\phi(x_i))$. Our modified objective function now becomes : 

\begin{figure*}
\begin{center}
 \includegraphics[width=0.24\linewidth]{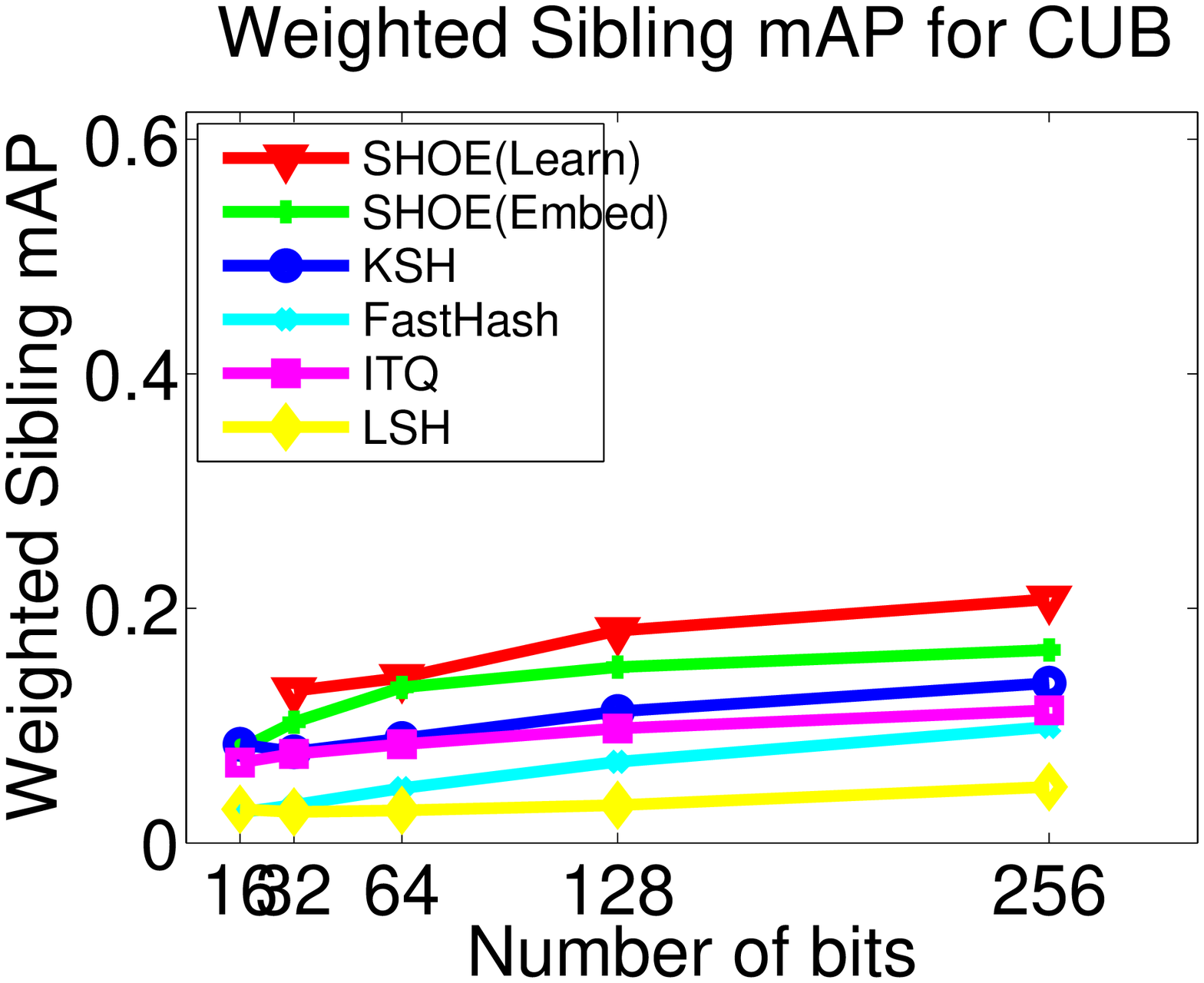}
 \includegraphics[width=0.24\linewidth]{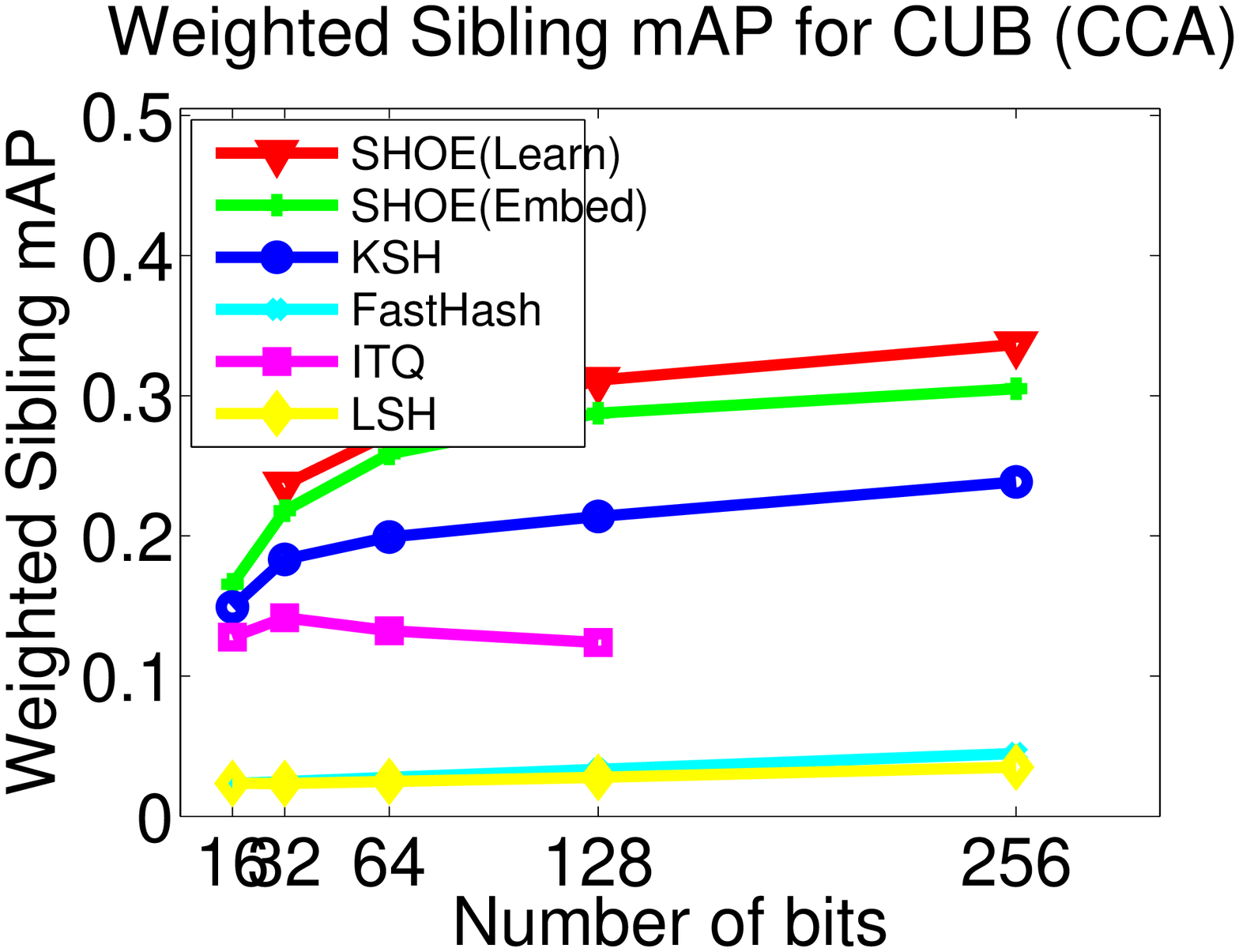}
 \includegraphics[width=0.24\linewidth]{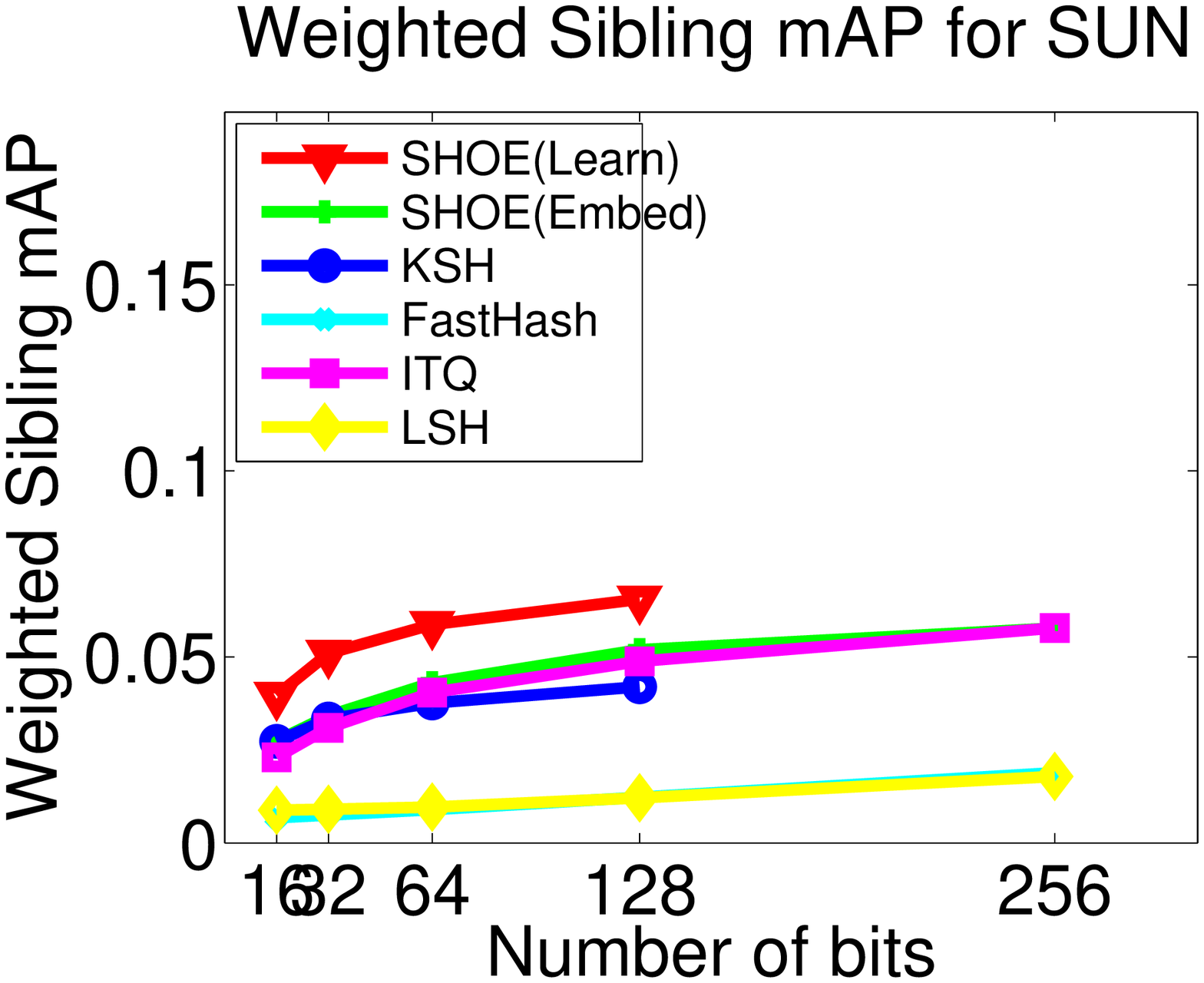}
 \includegraphics[width=0.24\linewidth]{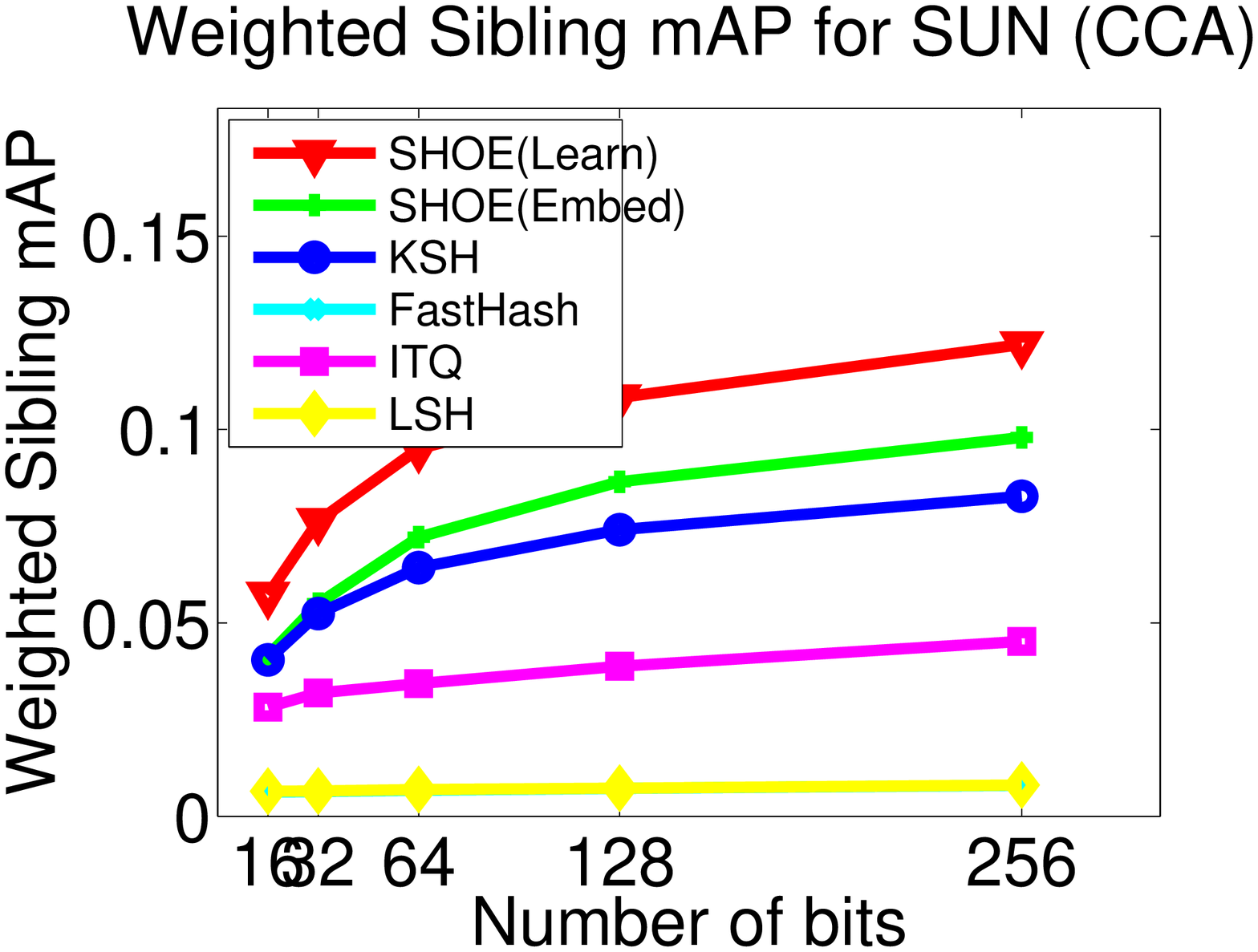}

 \includegraphics[width=0.24\linewidth]{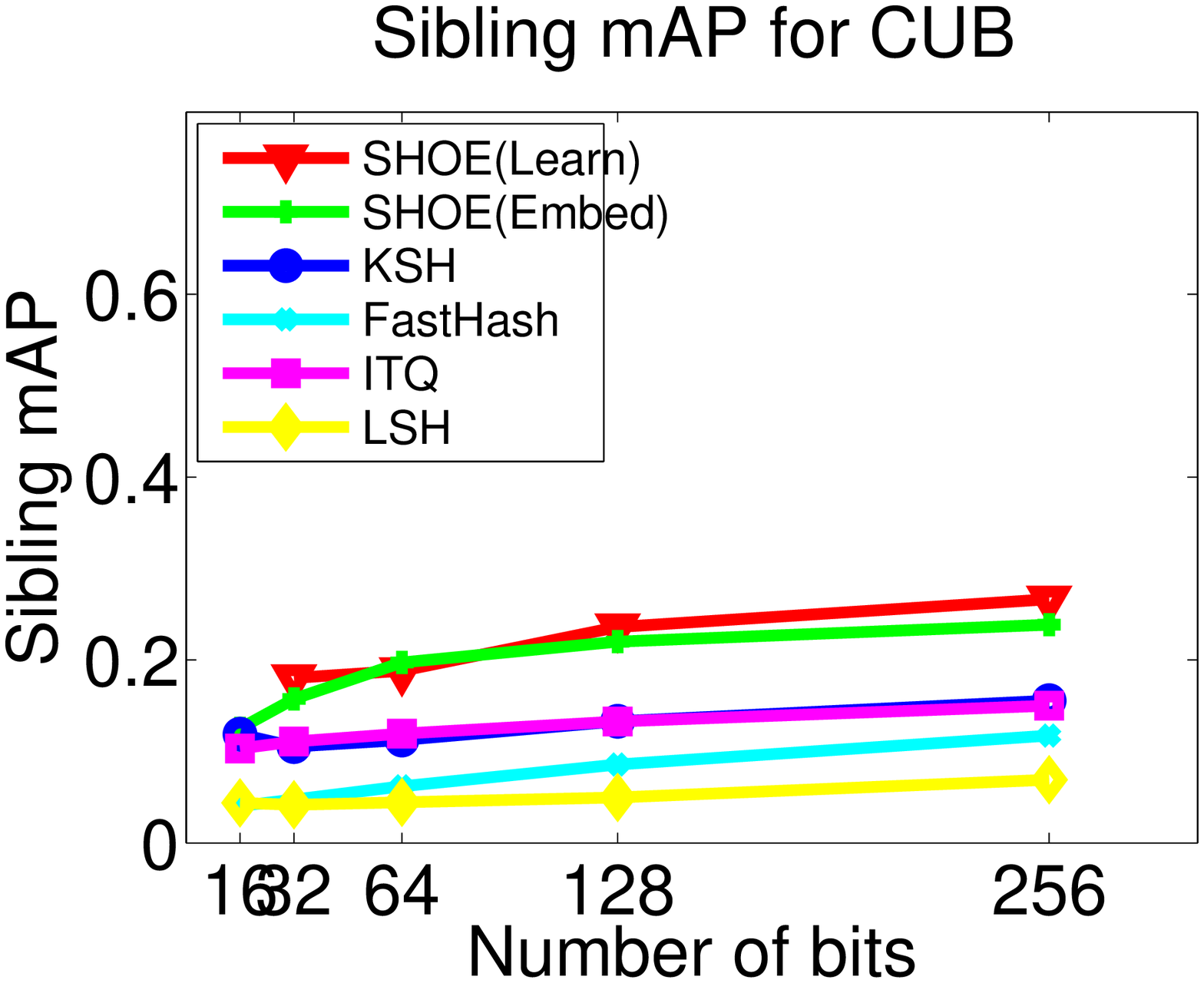}
 \includegraphics[width=0.24\linewidth]{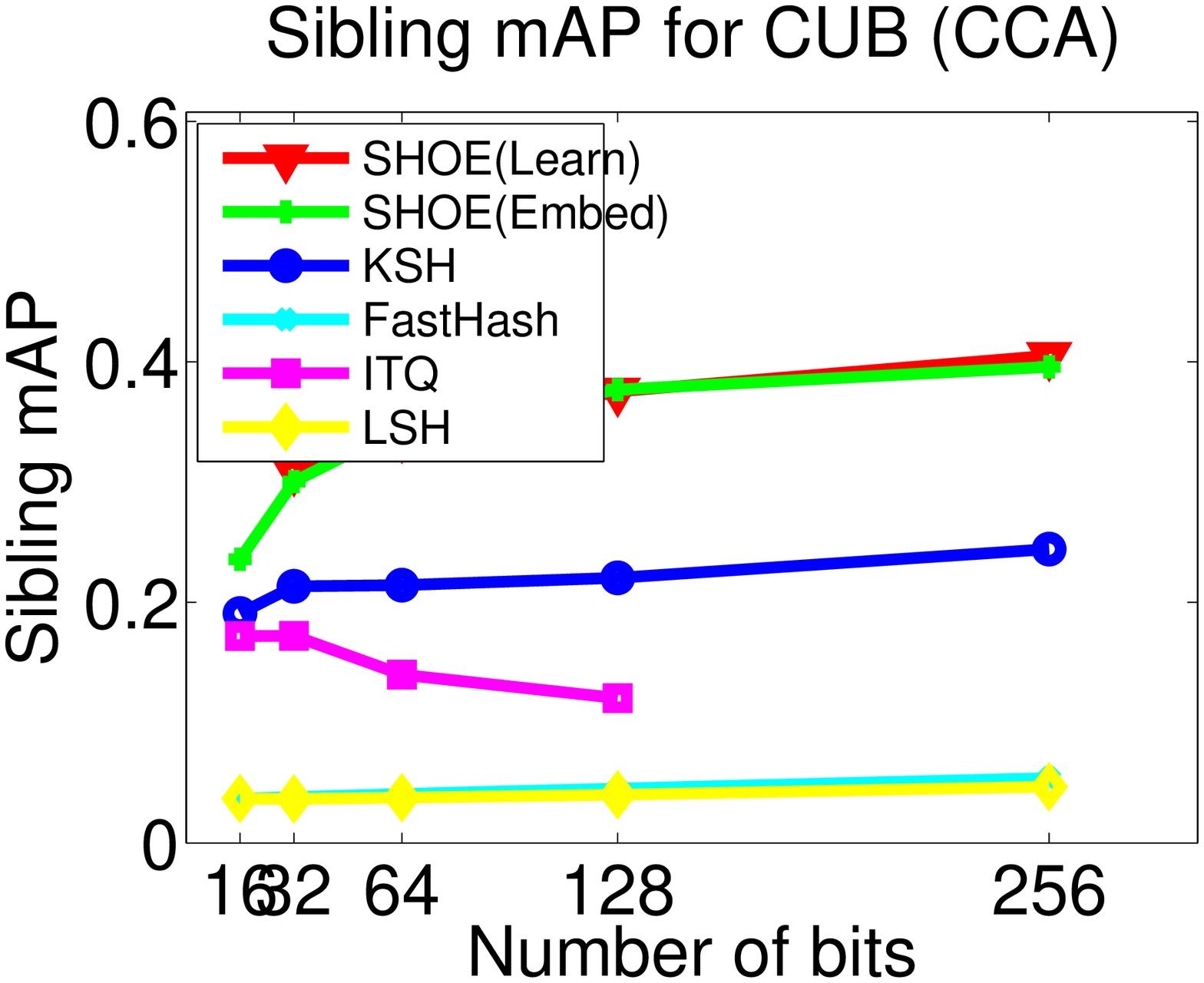}
 \includegraphics[width=0.24\linewidth]{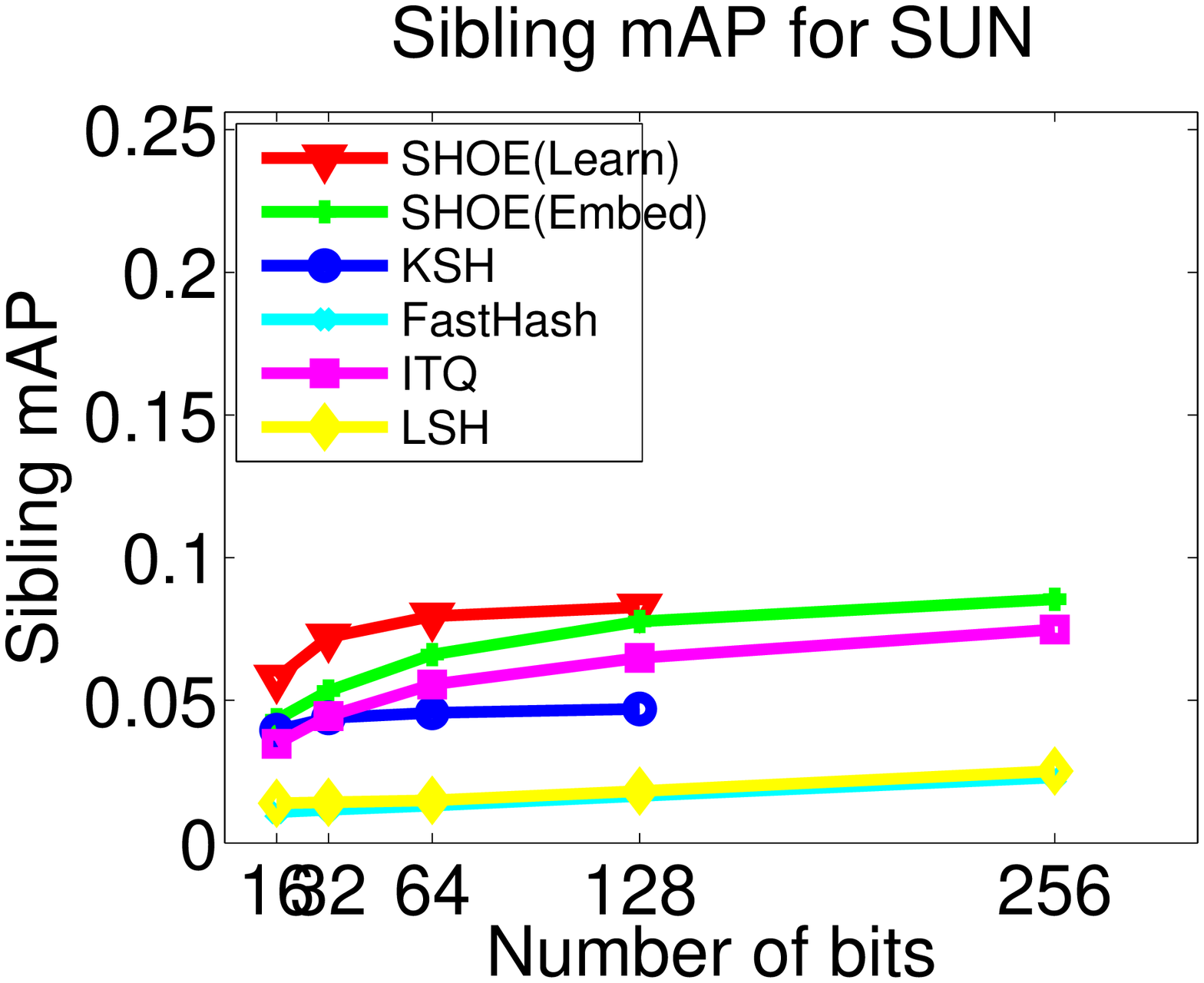}
 \includegraphics[width=0.24\linewidth]{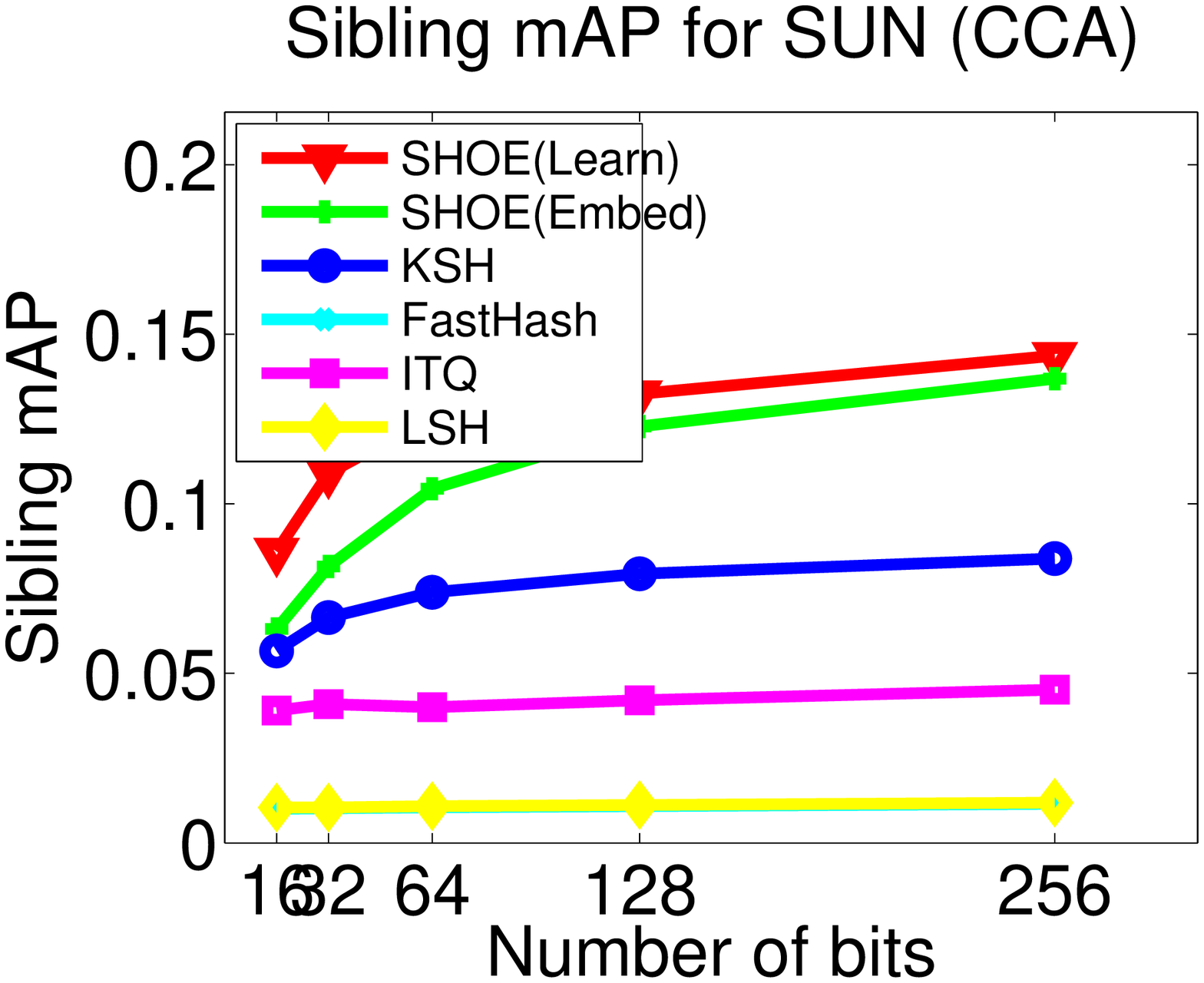}

 \includegraphics[width=0.24\linewidth]{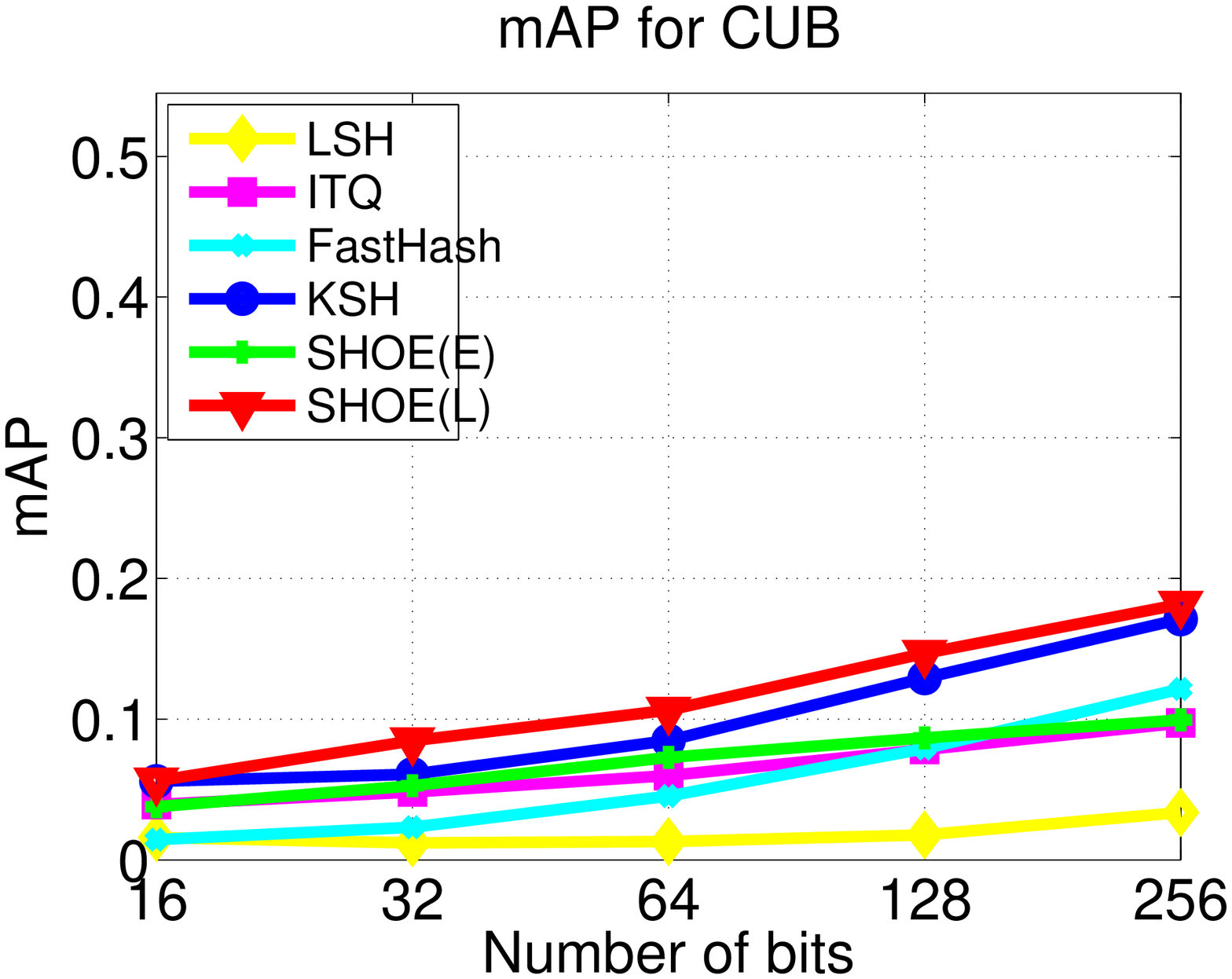}
 \includegraphics[width=0.24\linewidth]{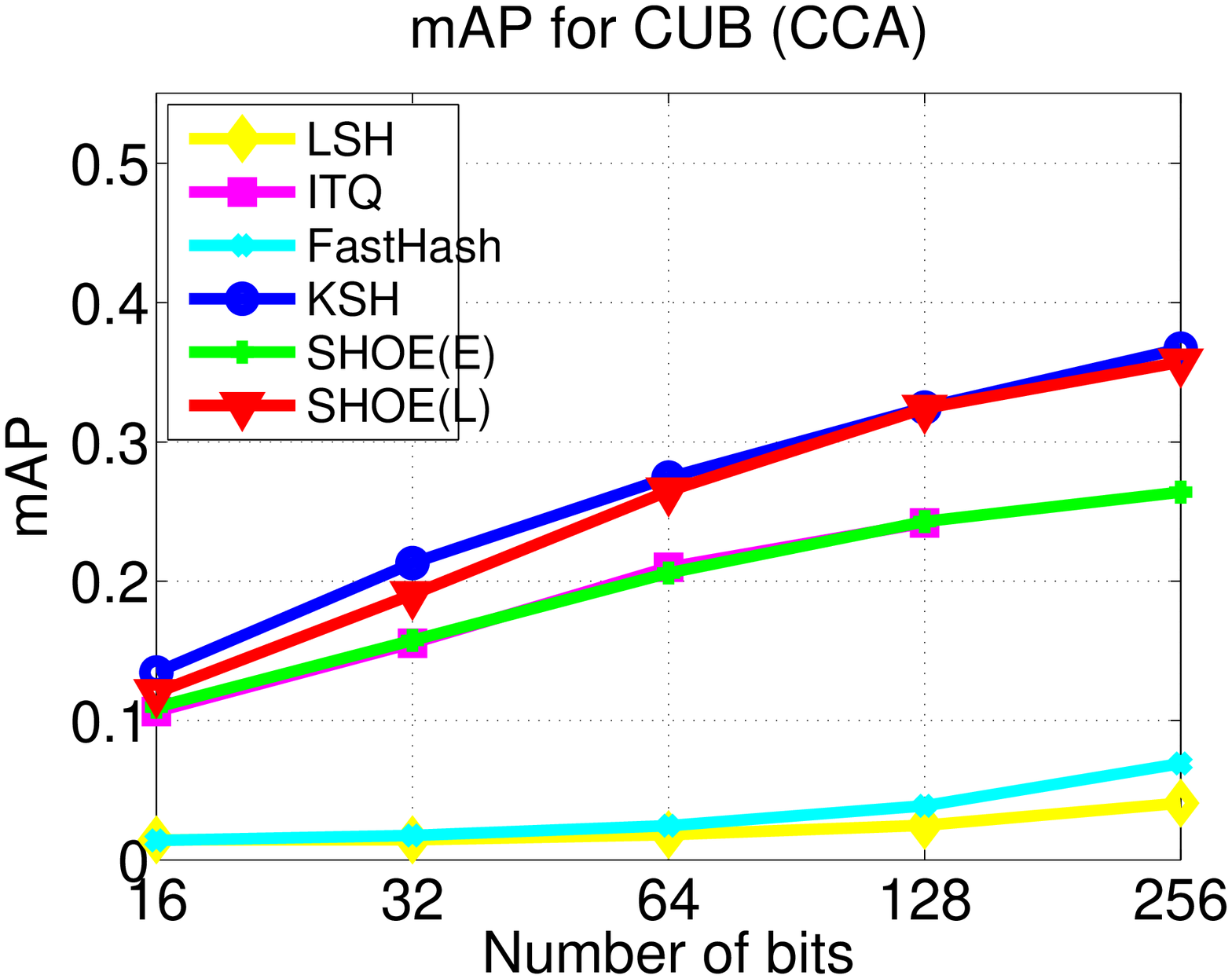}
 \includegraphics[width=0.24\linewidth]{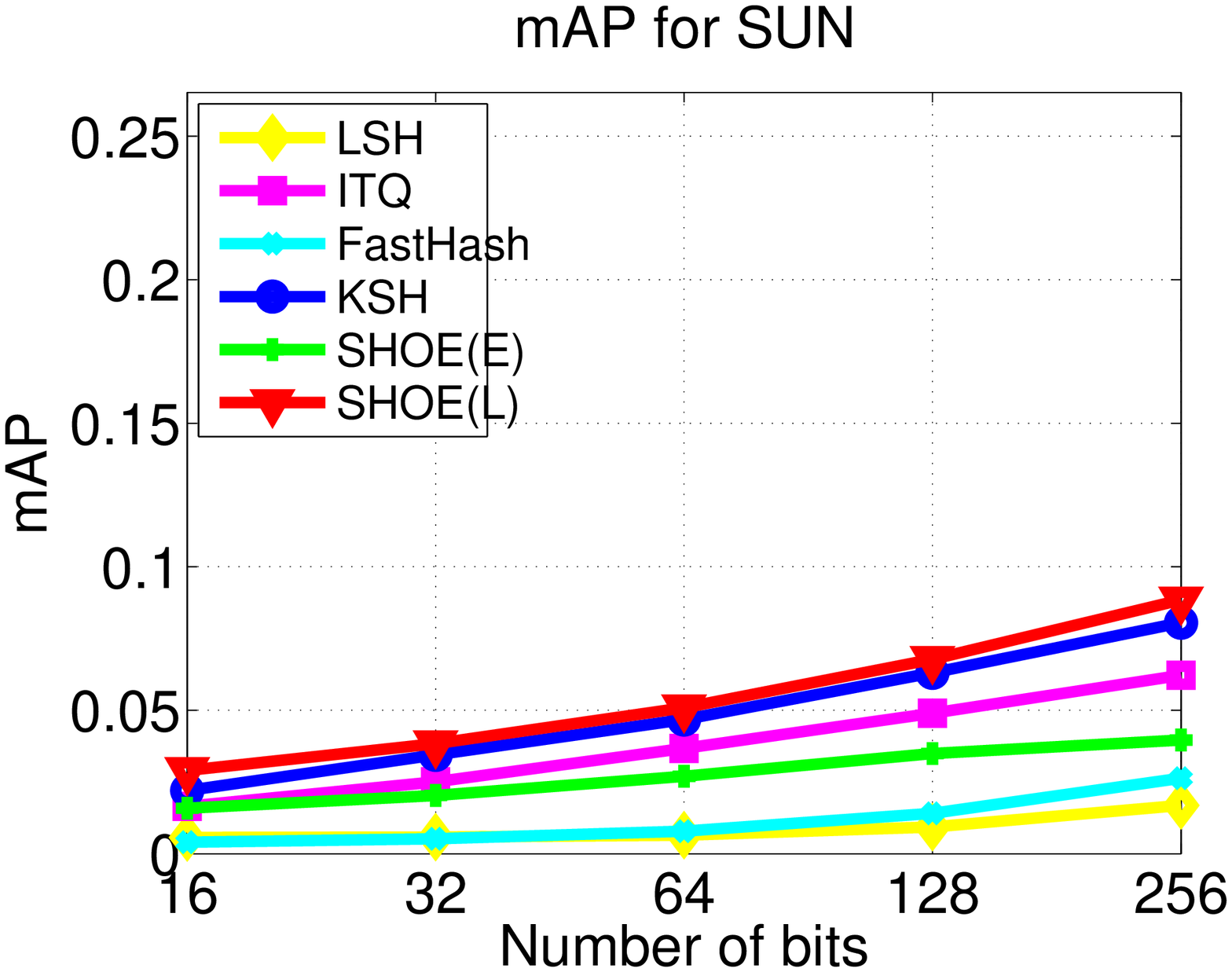}
 \includegraphics[width=0.24\linewidth]{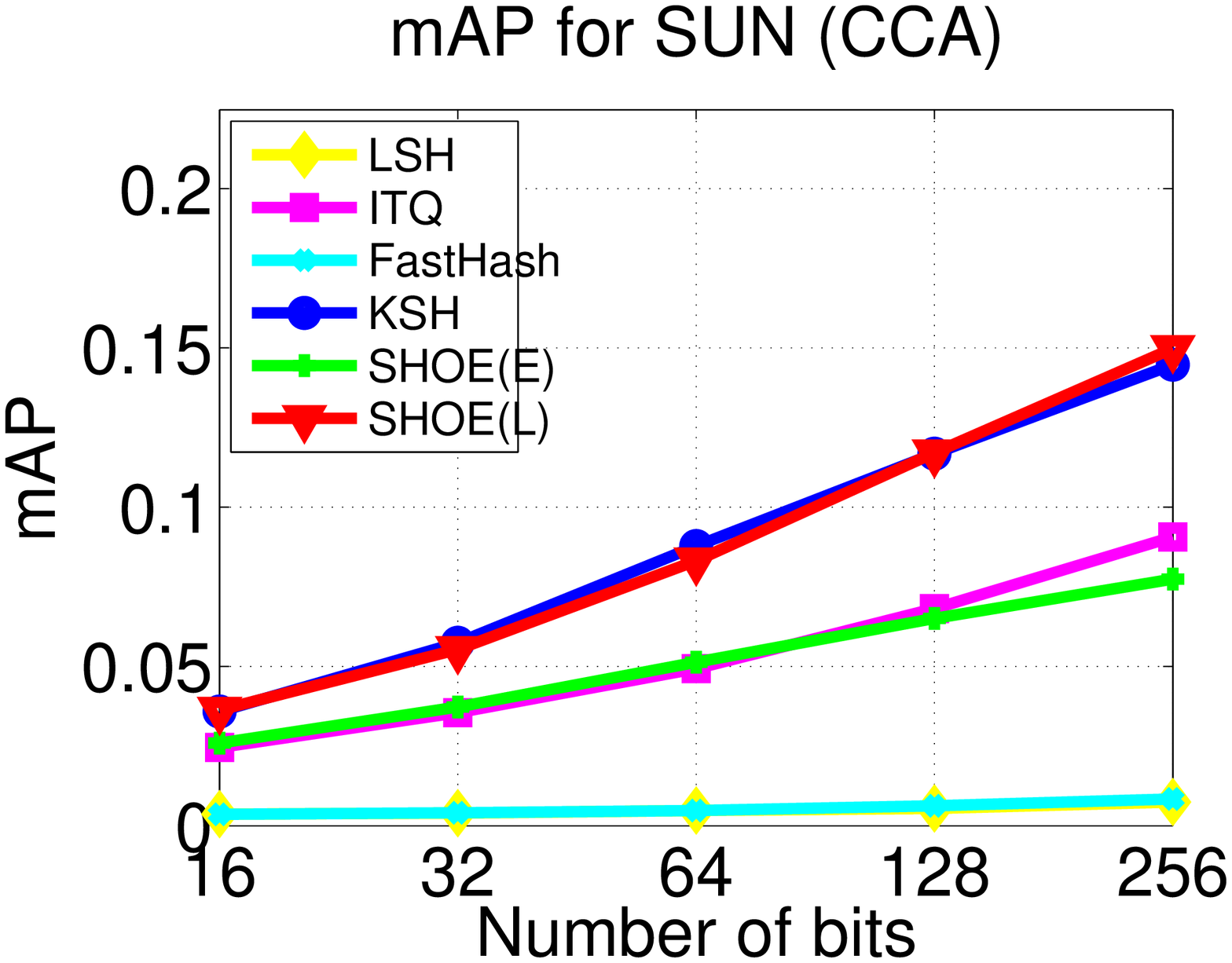}

\end{center}
   \caption{ Retrieval on CUB-2011(first and second column) and SUN(third and fourth column) dataset comparing our methods SHOE(E) and SHOE(L) with the state-of-the art hashing techniques. The above plots report mAP, Sibling and Weighted Sibling mAP for top 5 sibling classes.}
\label{fig:cubsunsmallsetlearn}
\end{figure*}


\small
\begin{gather}
   \min_{\substack{W, \theta}} \sum_{i,j \in}^N\sum_{\substack{same\\class}}^N ( \frac{1}{c}\sum_{l=1}^c h_{l_i} h_{l_j} - 1)^2
   + \sum_{i,j \in}^N\sum_{\substack{sibling\\class}}^N ( \frac{1}{c}\sum_{l=1}^c h_{l_i} h_{l_j} - \theta)^2 \nonumber\\
\label{eq:ourobj}   +\hspace{-0.05 cm}  \sum_{i,j \in}^N\sum_{\substack{unrelated\\class}}^N ( \frac{1}{c}\sum_{l=1}^c h_{l_i} h_{l_j} + 1)^2 + \lambda \norm{\theta + 1}^2 
\end{gather}

\normalsize
Let $\mathcal{H}_{{ij}_c} = \frac{1}{c}\displaystyle\sum_{l=1}^c h_{l_i} h_{l_j}$ denote the sum of the inner product of the binary codes $b_i$ and $b_j$ of length $c$. We now compute the derivative of Equation~\eqref{eq:ourobj} w.r.t $\theta$, set it to 0 and solve for $\theta$. We obtain :
\begin{equation} \label{eq:thetader}
   \theta = \theta_c = \frac{\displaystyle\sum_{i,j \in}^N\displaystyle\sum_{\substack{sibling\\class}}^N ( \mathcal{H}_{{ij}_c} - \lambda)}{c(n_{sib} + \lambda)}
\end{equation}

\noindent
where $n_{sib}$ is the number of sibling pairs in the training data. We have thus obtained a closed form solution for the optimal $\theta$. However, we cannot calculate $\theta$ directly as we do not learn all the bits at once. Therefore, we employ a two step alternate optimization procedure that first learns the bits and then an approximate $\theta_l$ value calculated from the previously learned bits. For the first iteration, we use an initial $\theta_0$ value, that satisfies the constraint: $-1<\theta_0<0$. The two step optimization procedure for learning the $l^{th}$ hash function is:
\begin{enumerate}
 \item Step 1 : We optimize for Equation~\eqref{eq:dotprodobj}, keeping $\theta_{l-1}$ constant and updating the projection vector $W$, thus learning hash-code bits $h_l(\phi(x_i))$.
 \item Step 2 : We keep the hash-code bits $h_l(\phi(x_i))$ constant and learn for $\theta_l$ using Equation~\eqref{eq:thetader}.
\end{enumerate}





\subsection{Supervised Dimensionality Reduction}
As our datasets contain class label information and corresponding output embeddings, we have explored the idea of 
supervised dimensionality reduction for input embeddings $\phi(x) \in R^d$ to $\omega(x) \in R^c (c \ll d) $, given the output embeddings $\psi(y) \in R^E$. There are many supervised dimensionality reduction techniques available in the literature like Canonical Correlation Analysis (CCA) \cite{cca} and Partial Least Square Regressions \cite{pls}, for example. In particular, we have used CCA \cite{cca} ($\phi \rightarrow \omega$) to extract a common latent space from two views that maximizes the correlation with each other. \cite{itq} also leveraged the label information using CCA to obtain supervised features prior to binary encoding. However, they limit their output embeddings to take the form of one vs remainder embeddings: : $\psi(y) \in \{0, 1\}^L$ is a $L$-dimensional binary vector with exactly one bit set to 1 i.e. $\psi(y)_y = 1$ where $L$ is the number of class labels.  On the other hand, we apply CCA to the general form of output embeddings that are real valued 
continuous attributes capturing structure between the classes. We observe that when supervised features with CCA-projections are used, we obtain a significant boost in performance ($\approx 100\%$ improvement) for all of our evaluation metrics.

\begin{table*}
\begin{center}
\begin{tabular}{|c|c|c|c|c|c|c|c|c|c|c|c|c|}
\hline
 Method & \multicolumn{6}{|c|}{CUB-2011(5000 train)} & \multicolumn{6}{|c|}{SUN Attribute(7000 train)} \\
\hline
\hline 
  & \multicolumn{2}{|c|}{$pre|mAP$} & \multicolumn{2}{|c|}{$Sib_{pre|mAP}$} & \multicolumn{2}{|c|}{$Sib^w_{pre|mAP}$} 
& \multicolumn{2}{|c|}{$pre|mAP$} & \multicolumn{2}{|c|}{$Sib_{pre|mAP}$} & \multicolumn{2}{|c|}{$Sib^w_{pre|mAP}$} \\
\hline
& @30 & mAP & @30 & mAP & @30 & mAP & @10 & mAP & @10 & mAP & @10 & mAP\\
\hline
SHOE(L)+CCA & 0.481 & \textbf{0.527}  &\textbf{0.701} & 0.529  & \textbf{0.617} & \textbf{0.436}  & 0.169 & 0.201  & \textbf{0.344} & \textbf{0.239}  & \textbf{0.269} & \textbf{0.193} \\
\hline
SHOE(E)+CCA & 0.387 & 0.429  & 0.668 & \textbf{0.533}  & 0.554 & 0.429  & 0.112 & 0.134  & 0.270 & 0.205  & 0.199 & 0.152 \\
\hline
KSH+CCA & \textbf{0.488} & 0.526  & 0.661 & 0.290  & 0.595 & 0.303  & \textbf{0.191} & \textbf{0.220}  & 0.300 & 0.130  & 0.252 & 0.126 \\
\hline
ITQ+CCA & 0.242 & 0.256  & 0.339 & 0.125  & 0.299 & 0.129  & 0.062 & 0.070  & 0.108 & 0.044  & 0.088 & 0.041 \\
\hline
FastHash & 0.240 & 0.246  & 0.341 & 0.120  & 0.298 & 0.124  & 0.021 & 0.021  & 0.038 & 0.017  & 0.030 & 0.014 \\
\hline
LSH & 0.024 & 0.017  & 0.082 & 0.049  & 0.054 & 0.032  & 0.010 & 0.009  & 0.034 & 0.018  & 0.022 & 0.012 \\
\hline
\end{tabular}
\end{center}
\caption{Comparing Precision, mAP and their sibling variants with our methods(SHOE(L) and SHOE(E)) and several baselines for 128 bits. We apply CCA projections for all the methods except FastHash and LSH as their performance decreases. For sibling metrics, SHOE performs significantly better than the baselines and performs as well as the baselines for standard metrics.}
\label{tab:prerecomparebig}
\end{table*}

\section{Experiments}
\label{sec:experiments}
We evaluate our method on the following datasets: Caltech-UCSD Birds (CUB) Dataset \cite{cub}, the SUN Attribute Dataset \cite{sun} and Imagenet ILSVRC2010 dataset \cite{ilsvrc2010}. We extract CNN features from \cite{caffe}, as mentioned in Section~\ref{sec:embedexp}. In the case of CUB dataset, we extract CNN features for the bounding boxes that accompany the images. For CUB and SUN datasets, we create two training sets of different size to examine the variation in performance with number of training examples. For all the datasets, we define the ground truth using class labels.

\textbf{\textit{Datasets:}} We have described the CUB dataset in Section 3. The ImageNet ILSVRC 2010 dataset is a subset of ImageNet and contains about 1.2 million images distributed amongst 1000 classes. We uniformly select 2 images per class as a test set and use the rest as retrieval set. We select 5000 training and $p=3000$ anchor point images by uniformly sampling across all classes. We obtain the output embeddings for the ImageNet class using the method of Tsochantaridis\cite{taxonomy}. Each of the 74401 synsets in ImageNet is a node in a hierarchy graph and using this graph, we obtain the ancestors for each class in ILSVRC 2010 dataset. We then construct a matrix $O_{Imagenet} = \{0,1\}_{1000 \times 74401}$, where 
the $j^{th}$ column of the $i^{th}$ row is set to 1 if the $j^{th}$ class is an ancestor of the $i^{th}$ class, 0 otherwise. Thus, the output embedding of each class is represented by a row of $O_{Imagenet}$.

The SUN Attribute dataset\cite{sun} contains 14340 images equally distributed amongst 717 classes, accompanied by annotations of 102 real valued attributes. We partition the dataset into equal retrieval and test sets, each containing 7170 images. We derive two variants from the retrieval set - the first has 3585(5 per class) training and 1434(2 per class) anchor point images, while the second has 7000(10 per class) training and 3000(~4 per class) anchor point images. In this dataset, each test query has 10 same class neighbors and 50 sibling class neighbors in the retrieval set. We compute a per class embedding by averaging embedding vectors for each image in the class.

\textbf{\textit{Evaluation Protocol:}} For binary hash-codes of length $c = \{16, 32, 64, 128, 256\}$, we evaluate SHOE using standard, sibling and weighted sibling flavors of precision, recall and mAP. Two variants of SHOE are used - SHOE(E), which uses raw output embeddings, and SHOE(L), which is learned using the method in Section \ref{subsec:ourmethod}. For the smaller CUB and SUN subsets, we compute mAP and their sibling versions. For the big subsets, in addition to mAP, we also compute precision. In the case of ImageNet, we compute precision@50, recall@10K, mAP and their sibling variants.


\textbf{\textit{Results:}}
We compare our method to the methods which are well known in image-hashing literature mentioned in section~\ref{sec:embedexp} and present the results in Figure~\ref{fig:cubsunsmallsetlearn} and Table~\ref{tab:prerecomparebig}.  We use CNN+K features as input embeddings for SHOE and KSH. For the other methods, we use only CNN features. The rows represent weighted sibling, sibling and mAP metrics. The first two columns represent experiments without and with CCA projections on the CUB dataset, while the latter two columns represent the same on the SUN Attribute dataset. From these plots, we see that our method SHOE(red and green), without CCA projections, comfortably outperforms the best baseline, KSH(blue) on all metrics for both datasets. Using the CCA projections, we significantly improve the performance(by $\approx 100\%$) for SHOE, KSH and ITQ, while it lowers the performance of FastHash and LSH. Our method with CCA projections performs as well as the KSH baseline on the mAP metric and outperforms on the sibling or weighted sibling metrics. The gap in performance between SHOE and KSH reduces when CCA projections are applied on the standard metrics, but not on the sibling metrics. On ImageNet, we only show results with CCA projected features. We observe a similar trend, with SHOE surpassing KSH on recall@10K and sibling precision metrics, while performing as well as KSH on precision@50 and mAP. 







%
\begin{figure*}
\begin{center}
\begin{minipage}[t]{.68\linewidth}
\vspace{0pt}
\centering
 \includegraphics[width=0.32\linewidth]{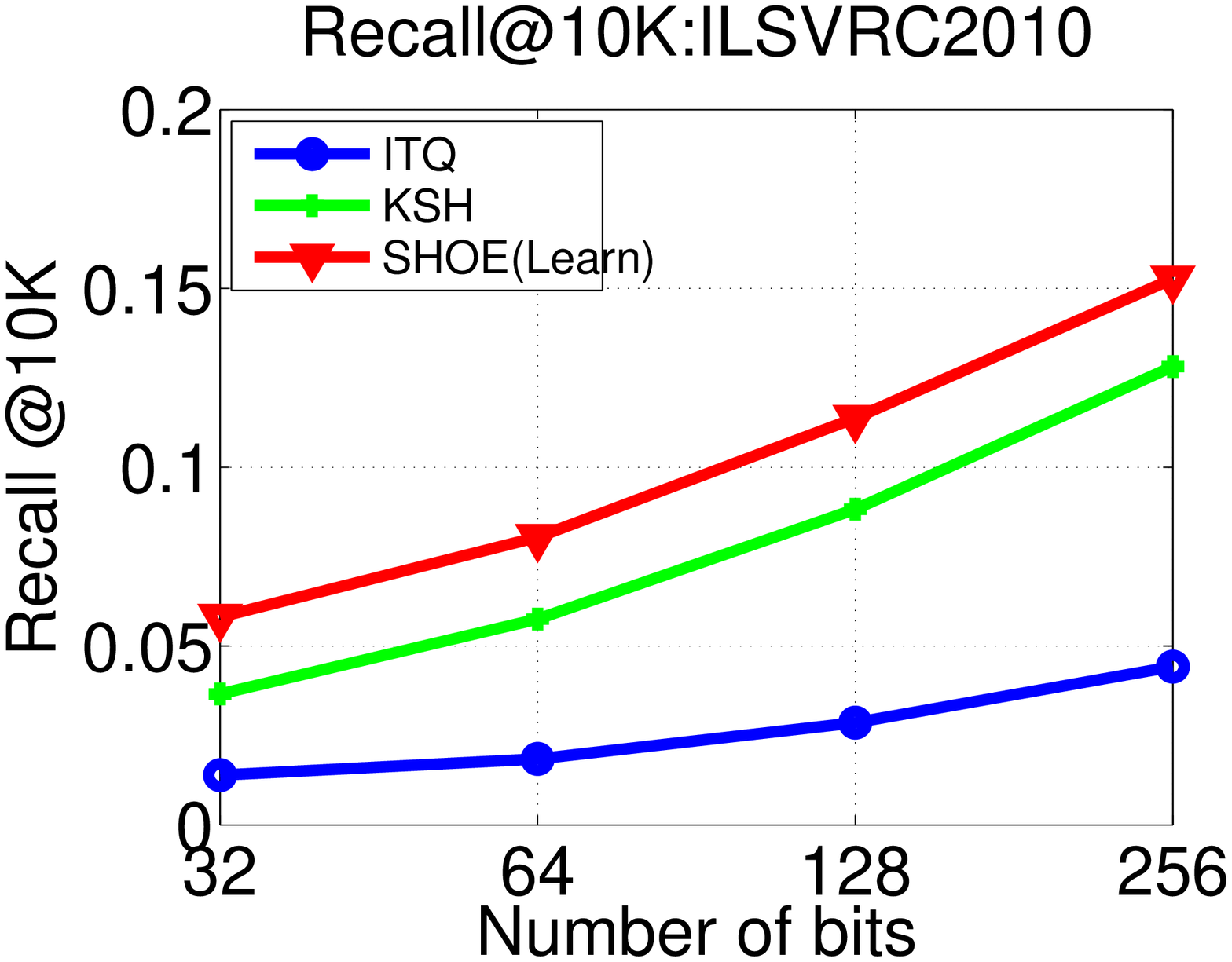}
 \includegraphics[width=0.32\linewidth]{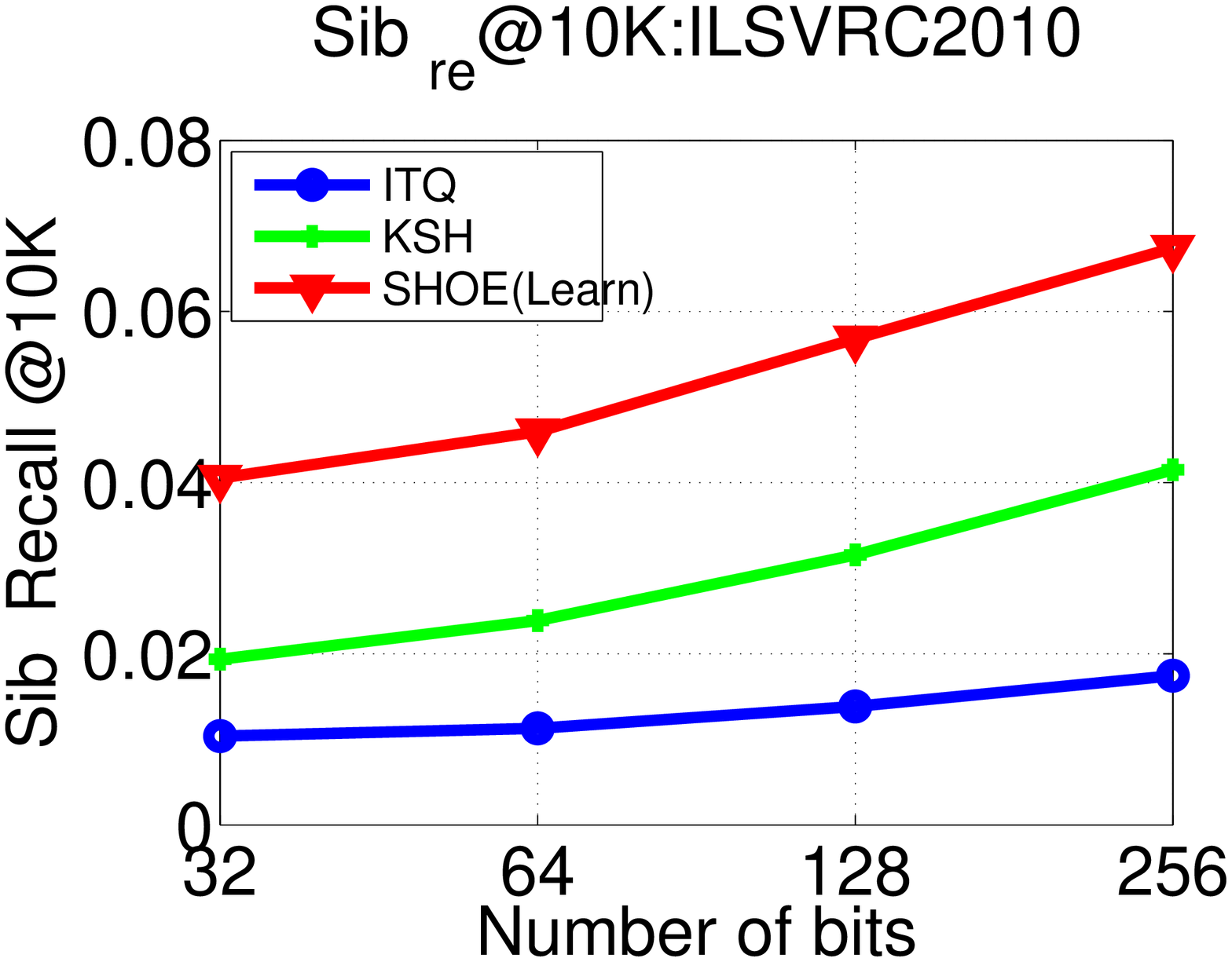}
 \includegraphics[width=0.32\linewidth]{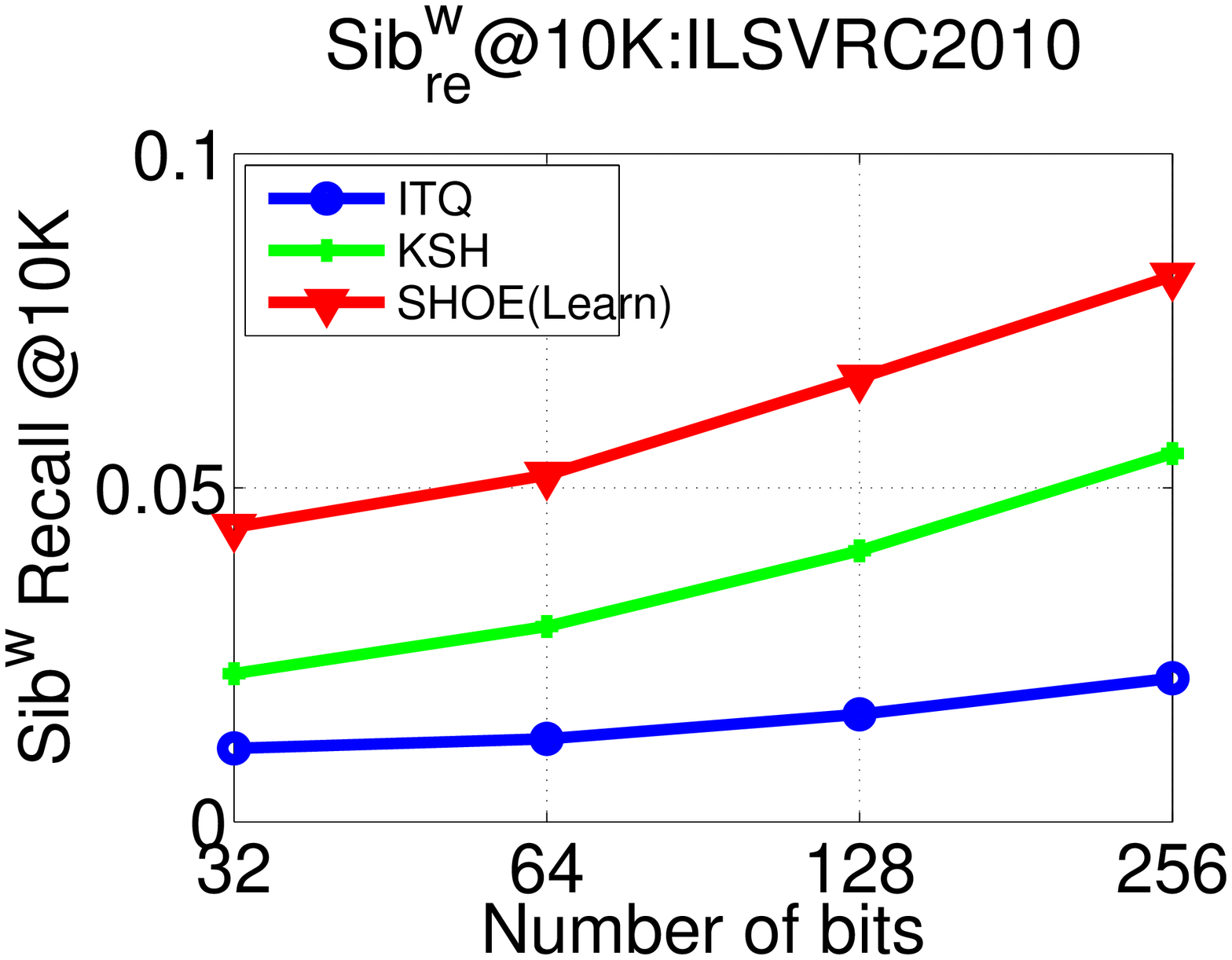}
\end{minipage}%
\begin{minipage}[t]{0.32\linewidth}
\vspace{0pt}
\centering
\small
\begin{tabular}{|c|c|c|c|}

\hline
Method & pre & $s_{pre}$ & $s_{pre}^w$  \\
\hline
SHOE(L)+CCA &  24.5 & \textbf{32.6} & \textbf{29.1}\\ \hline
KSH+CCA & \textbf{24.8}  & 30.4 & 28.0 \\ \hline
ITQ+CCA & 6.5  & 10.8 & 9.9 \\ \hline
\hline
Method & mAP & $s_{mAP}$ & $s_{mAP}^w$ \\ \hline
SHOE(L)+CCA &  \textbf{0.039} & \textbf{0.021} & \textbf{0.022}\\ \hline
KSH+CCA & 0.036  & 0.010 & 0.014 \\ \hline
ITQ+CCA & 0.005  & 0.001 & 0.002 \\ \hline
\end{tabular}
\end{minipage}
\normalsize
\end{center}
\caption{ Retrieval on ILSVRC2010 dataset comparing SHOE with state-of-the art hashing techniques. We use $5K$ training samples and CNN+K+CCA as features for all the binary encoding schemes. The above plots report $recall$, $Sib_{re}$, $Sib_{re}^w$ @10K for top 5 sibling classes for bits $c=\{32, 64, 128, 256\}$. Table reports precision@50, mAP and their sibling versions for $256$ bits. }
\label{fig:imagenetcompare}
\end{figure*}


\section{Fine-grained Category Classification}
\label{sec:classification}
In this section, we demonstrate the effectiveness of our proposed codes for fine-grained classification of bird categories in CUB-2011 dataset. We propose a simple nearest neighbor pooling classifier that classifies a given test image by assigning it to the most common label among the top-$k$ retrieved images. Let $R_x(q,x) \rightarrow rank $ give the ranking of the images retrieved based on our binary codes. Thus, $rank$ is 1 for the nearest neigbhor and $rank$ is $N$ for the farthest neighbor, where $N$ is the size of the database. Given such ranking, with $\mathcal{M}_q$ denoting the top-$k$ ranked neighbors of a new query $q$,  we define the k-nn pooling classifier as:
\begin{equation}
	class_{predict}(q) = \argmax_{\substack{y}} \sum_{x \in \mathcal{M}_q} \mathbb{I}(class(x)==y)
\end{equation}

We use the above model to obtain the classification accuracy on the CUB dataset with 200 categories from top-$10$ neighbors. In particular, we obtain the following accuracies: top-1 accuracy(top-1) measures if the predicted class matches the ground truth class, top-5 accuracy(top-5) measures if one of the top-5 predicted classes match the ground truth class and sibling accuracy(sib) measures if the predicted class is one of the sibling classes of the ground truth class.  As a baseline, we train a linear SVM model on the CNN features. We compare our proposed binary codes SHOE(L), KSH and state-of-the-art fine grained classification models that use CNN features. For this experiment, we use bounding box information, but do not use any part-based information available with the datasets. Hence, we do a fair comparison between methods with no part-based information. Table ~\ref{tab:classacc} shows the classification performance over the 5794 test images with approximately 30 images for each of the 200 categories. 

\begin{figure}[t]
\begin{center}
   \includegraphics[height=4.5cm,width=8.5cm]{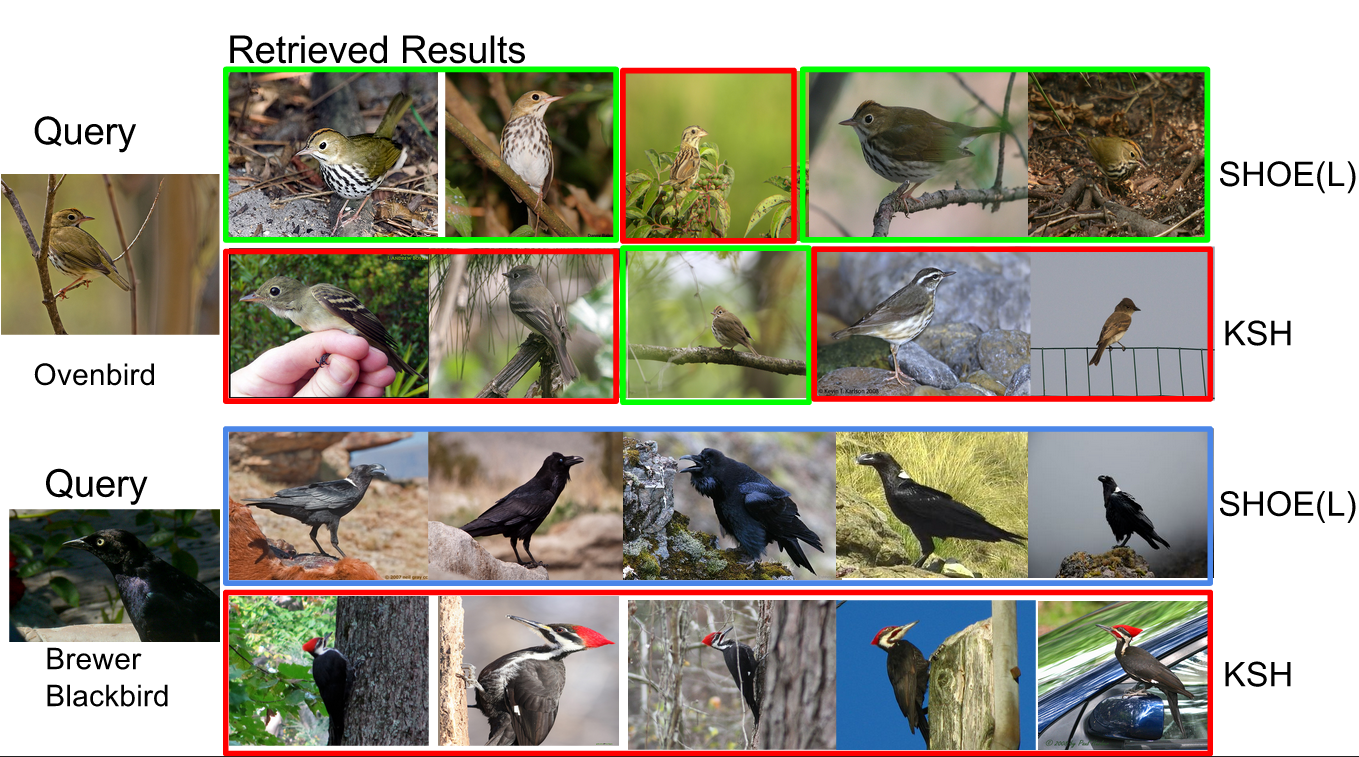}
\end{center}
   \caption{The first query is of an ovenbird. SHOE retrieves more ovenbirds than KSH. The second query is of a Brewer black-bird. Neither SHOE nor KSH retrieve Brewer black-birds. However, SHOE returns ravens, which are sibling classes of Brewer black-birds, whereas KSH retrieves pileated woodpeckers, which are unrelated to black-birds. Here, blue borders represent sibling classes. } 
\label{fig:cubqualres}
\end{figure}

\textbf{\textit{Features:}} For each of the binary coding schemes(SHOE, KSH, ITQ), we use the CNN+K+CCA(kernelized CNN with CCA projections) features as input embeddings and mean-centered attributes as the output embeddings. For the experiments, we used only 128 bit codes, while CNN features are 4096 dimensional vectors. 

\begin{table}
\begin{center}
\begin{tabular}{c|c|c|c|c}
\hline
 Method & top-1 & top-5 & sib & \small{Compression} \\
\hline
Baseline(SVM) & 50.6 & 75.6 & 70.19 & 1\\
\hline
\hline
SHOE(L)+CCA & \textbf{52.51} & \textbf{77.8} & \textbf{72.4} & \textbf{1024} \\
KSH+CCA & 52.48 & 75.1 & 69.06 & \textbf{1024}\\
ITQ+CCA & 27.5& 43.4 & 37.6& \textbf{1024}\\
R-CNN\cite{rcnn} & 51.5 & - & - & 1\\
Part-RCNN\cite{partrcnn} & 52.38 & - & - & 1\\
\hline
\end{tabular}
\end{center}
\caption{Comparing classification accuracies for CUB dataset. For top-1 and sibling accuracy, we used $k=10$ neighbors. To obtain top-5 accuracy, we used $k=50$ neighbors. For the binary coding schemes, we used only 5000 of the 5994 train images to obtain 128 bits, while the classification models are trained on the full set. '-' indicates that the information is not available in their paper.}
\label{tab:classacc}
\end{table}

\textbf{\textit{Results:}} We observe that not only do the proposed binary codes obtain a marginal improvement in performance over the complex classification models in \cite{rcnn} \cite{partrcnn}, but they also offer an astounding compression ratio of \textbf{1024}. Also, the training and testing times of binary coding schemes are significantly smaller than those with SVM classification models.

\section{Conclusion}
\label{sec:conclusion}
The key idea in our paper is to exploit output embeddings that capture relationships between classes and we use them to learn better hash functions for images. In our work, a method to learn class similarity jointly with the hash function was devised, along with new metrics for their evaluation. Our method SHOE, achieved state-of-the art image retrieval results over multiple datasets for hash codes of varying lengths. Our second innovation was to utilize CCA to learn a projection of features with output embeddings, which resulted in significant gains in both retrieval and classification experiments. Upon applying this approach to all methods, we perform as well or better than all baselines over all datasets. 

{\small
\bibliographystyle{ieee}
\bibliography{egbib}
}

\end{document}